\documentclass[letterpaper]{article}

\usepackage{arxiv}
\usepackage[utf8]{inputenc} 
\usepackage[T1]{fontenc}    
\usepackage{hyperref}       
\usepackage{url}            
\usepackage{booktabs}       
\usepackage{graphicx}
\usepackage{algpseudocode}
\usepackage{algorithm}
\usepackage{amsmath}
\usepackage{amssymb}
\usepackage{amsfonts}       
\usepackage{nicefrac}       
\usepackage{microtype}      
\usepackage{xcolor}         
\usepackage{xspace}
\usepackage[acronym]{glossaries}
\usepackage[inline]{enumitem}
\usepackage{subcaption}
\usepackage{chngcntr}
\usepackage{multirow}
\usepackage{tikz}
\usepackage[numbers]{natbib}
\usepackage[capitalize]{cleveref}

\newcommand{\eg}{e.g.\@\xspace}

\newcommand{\ie}{i.e.\@\xspace}
\newcommand{\etc}{etc...\@\xspace}
\DeclareMathOperator*{\argmax}{arg\,max}
\newcommand{\textbif}[1]{\textbf{\textsl{#1}}}
\newcommand{\Transpose}{\ensuremath{^\top}}
\newcommand{\Prob}[1]{\ensuremath{P\left(#1\right)}}

\newacronym{abode}{ABODe}{Annotated Behaviour and Observability Dataset}
\newacronym{alm}{ALM}{Activity Labelling Module}
\newacronym[longplural={Bounding Boxes},shortplural={BBoxes}]{bb}{BBox}{Bounding Box}
\newacronym{bc}{BC}{Behaviour Classifier}
\newacronym{btp}{BTI}{Behaviour Time Interval}
\newacronym{em}{EM}{Expectation Maximisation}
\newacronym{fn}{FN}{False Negative}
\newacronym{fp}{FP}{False Positive}
\newacronym{gbm}{GBM}{Global Behaviour Model}
\newacronym{hmm}{HMM}{Hidden Markov Model}
\newacronym{imadge}{IMADGE}{Individual Mouse Activity Dataset for Group Environments}
\newacronym{lfb}{LFB}{Long-term Feature Bank}
\newacronym{lgr}{LgR}{Logistic Regression}
\newacronym{mnap}{mAP}{Mean Average-Precision}
\newacronym{mrc}{MLC at MRC Harwell}{Mary Lyon Centre at MRC Harwell, Oxfordshire}
\newacronym{nb}{NB}{Na\"\i ve Bayes}
\newacronym{oc}{OC}{Observability Classifier}
\newacronym{pca}{PCA}{Principal Component Analysis}
\newacronym{rdl}{RDL}{Relative Difference in Log-Likelihood}
\newacronym{rfid}{RFID}{Radio-Frequency Identification}
\newacronym{sgd}{SGD}{Stochastic Gradient Descent}
\newacronym{stlt}{STLT}{Spatio-Temporal Localisation Transformer}
\newacronym{tim}{TIM}{Tracking and Identification Module}
\newacronym{tn}{TN}{True Negative}
\newacronym{tp}{TP}{True Positive}

\newcommand{\NOOB}{\texttt{Not Observable}\@\xspace}
\newcommand{\OBS}{\texttt{Observable}\@\xspace}
\newcommand{\PQ}{\ensuremath{\mathbf{\xi}}}
\newcommand{\PX}{\ensuremath{\Psi}}

\newcommand{\PZ}{\ensuremath{\mathbf{\pi}}}
\newcommand{\FS}{F\textsubscript{1}\@\xspace}
\newcommand{\strain}{C57BL/6NTac\@\xspace}
\newcommand{\SData}{\ensuremath{\mathcal{D}}}
\newcommand{\TZ}{\ensuremath{\Omega}}
\newcommand{\urlabode}{\url{https://github.com/michael-camilleri/ABODe}}
\newcommand{\urlimadge}{\url{https://github.com/michael-camilleri/IMADGE}}
\newcommand{\IMM}{\texttt{Immobile}\@\xspace}
\newcommand{\FEED}{\texttt{Feeding}\@\xspace}
\newcommand{\DRINK}{\texttt{Drinking}\@\xspace}
\newcommand{\SGRM}{\texttt{Self-Grooming}\@\xspace}
\newcommand{\AGRM}{\texttt{Allo-Grooming}\@\xspace}
\newcommand{\LOCO}{\texttt{Locomotion}\@\xspace}
\newcommand{\OTR}{\texttt{Other}\@\xspace}
\newcommand{\ALLLBLS}[1]{\IMM, \FEED, \DRINK, \SGRM, \AGRM, \LOCO #1 \OTR}
\newcommand{\UNREL}{\emph{Unreliable}\@\xspace}
\newcommand{\WASTE}{\emph{Wasteful}\@\xspace}
\newcommand{\nll}{\ensuremath{\widehat{\mathcal{L}}}}
\newcommand{\SParams}{\ensuremath{\Theta}}
\newcommand{\logprob}{\log\left(\Prob{\SParams;\mathcal{A}}\right)}
\newcommand{\QFunc}{\ensuremath{\mathcal{Q}}}
\newcommand{\HID}{\texttt{Hidden}\@\xspace}
\newcommand{\NOID}{\texttt{Unidentifiable}\@\xspace}
\newcommand{\TENT}{\texttt{Tentative}\@\xspace}
\newcommand{\CLIMB}{\texttt{Climbing}\@\xspace}
\newcommand{\UMVE}{\texttt{Micro-motion}\@\xspace}
\newcommand{\AMB}{\texttt{Ambiguous}\@\xspace}
\newcommand{\PHENO}{phenotyper\@\xspace}
\newcommand{\CODE}{\url{https://github.com/michael-camilleri/Mice-N-Mates}}

\glsdisablehyper
\crefname{section}{Sec.}{Secs.}
\Crefname{section}{Section}{Sections}
\Crefname{table}{Table}{Tables}
\crefname{table}{Tab.}{Tabs.}

\renewcommand{\shorttitle}{Of Mice and Mates}
\renewcommand{\headeright}{Published in \emph{IJCV (2024)}}
\renewcommand{\undertitle}{}

\begin{document}


\title{
\makebox[0pt][l]{\raisebox{1in}[0pt][0pt]{\hspace{-0.75in} \parbox{9in}{\small Accepted manuscript version of the paper published in \emph{IJCV} 2024 (DOI: \texttt{10.1007/s11263-024-02118-3})}}}
Of Mice and Mates: Automated Classification and Modelling of Mouse Behaviour in Groups using a Single Model across Cages
}

\author{%
Michael P. J. Camilleri\\
School of Informatics\\
University of Edinburgh\\
Edinburgh\\
\texttt{\small michael.p.camilleri@ed.ac.uk}
\And
Rasneer S. Bains\\
Mary Lyon Centre\\
MRC Harwell\\
Oxfordshire\\
\texttt{\small r.bains@har.mrc.ac.uk}
\And
Christopher K. I. Williams\\
School of Informatics\\
University of Edinburgh\\
Edinburgh\\
\texttt{\small ckiw@inf.ed.ac.uk}
}

\maketitle

\begin{abstract}
Behavioural experiments often happen in specialised arenas, but this may confound the analysis.
To address this issue, we provide tools to study mice in the home-cage environment, equipping biologists with the possibility to capture the temporal aspect of the individual's behaviour and model the interaction and interdependence between cage-mates with minimal human intervention.
Our main contribution is the novel \glsfirst{gbm} which summarises the joint behaviour of groups of mice across cages, using a permutation matrix to match the mouse identities in each cage to the model.
In support of the above, we also (a) developed the \glsfirst{alm} to automatically classify mouse behaviour from video, and (b) released two datasets, \acrshort{abode} for training behaviour classifiers and \acrshort{imadge} for modelling behaviour.
\end{abstract}




\section{Introduction}\label{S_INTRO}
Understanding behaviour is a key aspect of biology, psychology and social science, \eg for studying the effects of treatments~\citep{CBD_059}, the impact of social factors~\citep{CBD_030} or the link with genetics~\citep{PHT_006}.
Biologists often turn to model organisms as stand-ins, of which mice are a popular example, on account of their similarity to humans in genetics, anatomy and physiology~\citep{CBD_033}.
Traditionally, biological studies on mice have taken place in carefully controlled experimental conditions~\citep{CBD_033}, in which individuals are removed from their home-cage, introduced into a specific arena and their response to stimuli (\eg other mice) investigated: see \eg the work of \cite{CBD_024, CBD_052, CBD_036, CBD_041, CBD_038, VL_022, VL_023}.
This is attractive because:
(a) it presents a controlled stimulus-response scenario that can be readily quantified~\citep{PHT_017}, \textsl{and}
(b) it lends itself easier to automated means of behaviour quantification \eg through top-mounted cameras in a clutter-free environment~\citep{VL_022, CBD_024, CBD_041, CBD_038}.

The downside of such `sterile' environments is that they fail to take into account all the nuances in their behaviour~\citep{CBD_055}.
Such stimuli-response scenarios presume a simple forward process of perception-action which is an over-simplification of their agency~\citep{CBD_055}.
Moreover, mice are highly social creatures, and isolating them for specific experiments is stressful and may confound the analysis \citep{CBD_026, CBD_037}.
For these reasons, research groups, such as the International Mouse Phenotype Consortium \citep{PHT_013} and TEATIME cost-action\footnote{\url{https://www.cost-teatime.org}} amongst others, are advocating for the long-term analysis of rodent behaviour in the home-cage.
This is aided by the proliferation of home-cage monitoring systems, but is hampered by the shortage of automated means of analysis.

In this work, we tackle the problem of studying mice in the home-cage, giving biologists tools to analyse the temporal aspect of an individual's behaviour and model the interaction between cage-mates --- while minimising disruption due to human intervention.
Our contributions are:
(a) a novel \gls{gbm} for detecting patterns of behaviour in a group setting across cages,
(b) the \gls{alm}, an automated pipeline for inferring mouse behaviours in the home-cage from video, \textsl{and} 
(c) two datasets, \acrshort{abode} for automated activity classification and \acrshort{imadge} for analysis of mouse behaviours, both of which we make publicly available.

In this paper, we first introduce the reader to the relevant literature in \cref{S_RELATED}.
\Cref{S_DATASETS} describes the nature of our data, including the curation of two publicly available datasets: this allows us to motivate the methods which are detailed in \cref{S_METHODS}.
We continue by describing the experiments during model fitting and evaluation in \cref{S_EXP} and conclude with a discussion of future work (\cref{S_CONC}).


\section{Related Work}\label{S_RELATED}

\subsection{Experimental Setups}\label{SS_REL_STPS}
Animal behaviour has typically been studied over short periods in specially designated arenas --- see \eg~\cite{CBD_024, CBD_036, CBD_052} --- and under specific stimulus-response conditions~\citep{CBD_041}.
This simplifies data collection, but may impact behaviour~\citep{PHT_019} and is not suited to the kind of long-term studies in which we are interested.
Instead, newer research uses either an enriched cage~\citep{VL_023, VL_021, VL_031, MS_001} or, as in our case, the home-cage itself~\citep{CBD_026, VL_034}.
The significance of the use of the home-cage cannot be overstated.
It allows for capturing a wider plethora of nuanced behaviours with minimal intervention and disruption to the animals, but it also presents greater challenges for the automation of the analysis, and indeed, none of the systems we surveyed perform \emph{automated behaviour classification} for \emph{individual} mice in a \emph{group-housed} setting.

Concerning the number of observed individuals, single-mice experiments are often preferred as they are easier to phenotype and control~\citep{CBD_038, VL_023, VL_021, MS_001}.
However, mice are highly social creatures and isolating them affects their behaviour~\citep{CBD_037}, as does handling (often requiring lengthy adjustment periods).
Obviously, when modelling social dynamics, the observations must perforce include multiple individuals.
Despite this, there are no automated systems that consider the behaviour of each individual in the home-cage as we do.
Most research is interested in the behaviour of the group as a whole~\citep{CBD_036, VL_035, DS_007, VL_081}, which circumvents the need to identify the individuals.
\cite{CBD_034} do model a group setting, but focus on the mother only and how it relates to its litter: similarly, the social interaction test~\citep{CBD_036, VL_056} looks at the social dynamics, but only from the point of view of a resident/intruder and in a controlled setting.
While \cite{VL_032, VL_034} and \cite{VL_095} do model interactions, their setup is considerably different in that (a) they use specially-built arenas (not the home-cage), (b) use a top-mounted camera (which is not possible in the home-cage) and (c) classify positional interactions (\eg Nose-to-Nose, Head-to-Tail \etc, based on fixed proximity/pose heuristics) and not the type of individual activity (\eg Feeding, Drinking, Grooming \etc).

\subsection{Automated Behaviour Classification}\label{SS_REL_BC}
Classifying animal behaviour has lagged behind that of humans, with even recent work using manual labels~\citep{CBD_052, CBD_034, CBD_060}.
Automated methods often require heavy data engineering~\citep{VL_002, CBD_036, VL_081, VL_023}.
Animal behaviour inference tends to be harder because human actions are more recognisable~\citep{VL_031}, videos are usually less cluttered~\citep{VL_036} and most challenges in the human domain focus on classifying short videos rather than long-running recordings as in animal observation~\citep{VL_081}.
Another factor is the limited number of publicly available animal observation datasets that target the home-cage.
Most --- RatSI~\citep{DS_006}, MouseAcademy~\citep{CBD_041}, CRIM13~\citep{DS_007}, MARS~\citep{VL_056}, PDMB~\citep{VL_081}, CalMS21~\citep{Sun2021TheMB} and MABe22~\citep{pmlr-v202-sun23g} --- use a top-mounted camera in an open field environment: in contrast, our side-view recording of the home-cage represents a much more difficult viewpoint with significantly more clutter and occlusion.
Moreover, PDMB only considers pose information, while CRIM13, MARS and CalMS21 deal exclusively with a resident-intruder setup, focusing on global interactions between the two mice rather than individual actions.
We aim, by releasing \acrshort{abode} (\cref{SS_DATA_ABODE}), to fill this gap.

\subsection{Modelling Mouse Behaviour}
The most common form of behaviour analysis involves reporting summary statistics: \eg of the activity levels~\citep{VL_050}, the total duration in each behaviour~\citep{VL_034} or the number of bouts~\citep{VL_056}, effectively throwing away the temporal information.
Even where temporal models are used as by~\cite{CBD_036}, this is purely as an aid to the behaviour classification with statistics being still reported in terms of total duration in each state (behaviour).
This approach provides an incomplete picture, and one that may miss subtle differences~\citep{CBD_001} between individuals/groups.
Some research output does report ethograms of the activities/behaviours through time \citep{PHT_006, VL_056, VL_087} --- and \cite{CBD_026} in particular model this through sinusoidal functions --- but none of the works we surveyed consider the temporal co-occurrence of behaviours between individuals in the cage as we do.
For example, in MABe22, although up to three mice are present, the four behaviour labels are a multi-label setup, which indicate whether each action is evidenced at each point in time, but not which of the mice is the actor.
This limits the nature of the analysis as it cannot capture inter individual dynamics, which is where our analysis comes in.

An interesting problem that emerges in biological communities is determining whether there is evidence of different behavioural characteristics among individuals/groups~\citep{CBD_060, CBD_033, CBD_034} or across experimental conditions~\citep{CBD_001, CBD_026}.
Within the statistics and machine learning communities, this is typically the domain of \emph{anomaly detection} for which \cite{AMD_034} provide an exhaustive review.
This is at the core of most biological studies and takes the form of hypothesis testing for significance~\citep{CBD_034}.
The limiting factor is often the nature of the observations employed, with most studies based on frequency (time spent or counts) of specific behaviours~\citep{CBD_037, CBD_062}.
The analysis by~\cite{CBD_034} uses a more holistic temporal viewpoint, albeit only on individual mice, while our models consider multiple individuals.
\cite{CBD_038} employ \glspl{hmm} to identify prototypical behaviour (which they compare across environmental and genetic conditions) but only consider pose features --- body shape and velocity --- and do so only for individual mice.
To our knowledge, we are the first to use a global temporal model inferred across cages to flag `abnormalities' in another demographic.


\section{Datasets} \label{S_DATASETS}

\begin{figure*}
\centering
	\includegraphics[width=\textwidth]{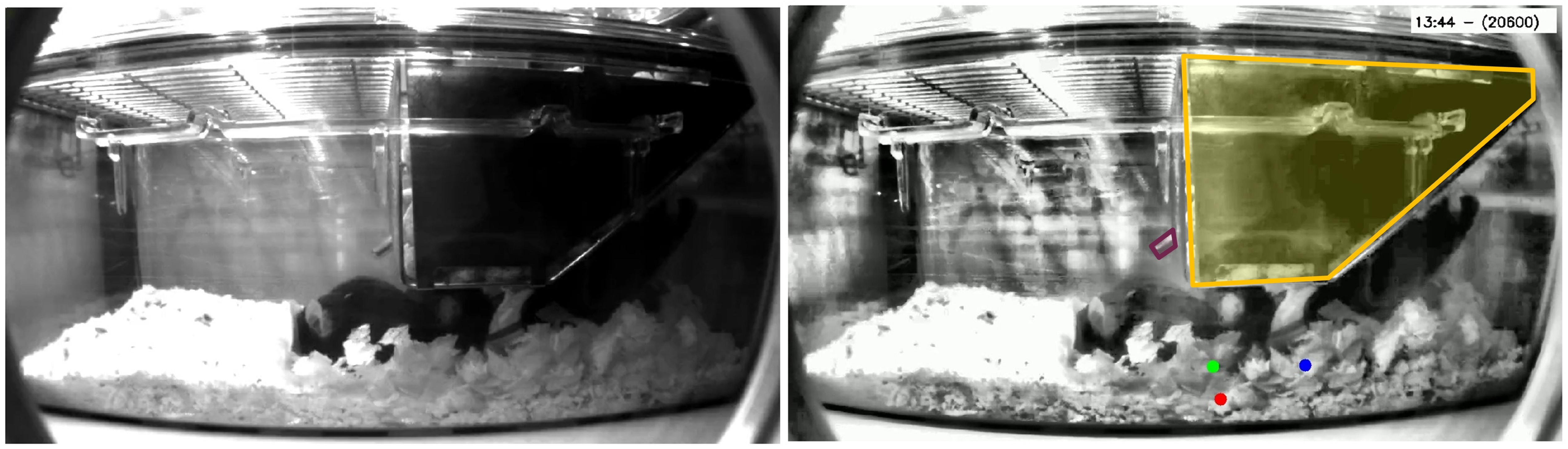}
	\caption{
	An example video frame from our data, showing the raw video (left) and an enhanced visual (right) using CLAHE \citep{MISC_038}.
	In the latter, the hopper is marked in yellow and the water spout in purple, while the (\acrshort{rfid}) mouse positions are projected into image space and overlaid as red, green and blue dots.
	}
	\label{FIG_DATA_FRAME}
\end{figure*}

A key novelty of this work relates to the use of continuous recordings of \emph{group-housed mice in the home-cage}.
In line with the Reduction strategy of the 3Rs \citep{PHT_022} we reuse existing data already recorded at the \gls{mrc}.
In what follows, we describe the modalities of the data (\cref{SS_DATA_MODALITIES}), documenting the opportunities and challenges this presents, as well as our efforts in curating and releasing two datasets to solve the behaviour modelling (\cref{SS_DATA_IMADGE}) and classification (\cref{SS_DATA_ABODE}) tasks.

\subsection{Data Sources}\label{SS_DATA_MODALITIES}
We use continuous three-day video and position recordings --- captured using the home-cage analyses system of \citet{CBD_026} --- of group-housed mice of the same sex (male) and strain (\strain).

\subsubsection{Husbandry}
The mice are housed in groups of three as a unique cage throughout their lifetime.
To reduce the possibility of impacting social behaviour \citep{CBD_055}, the mice have no distinguishing external visual markings: instead, they are microchipped with unique \gls{rfid} tags placed in the lower part of their abdomen.
All recordings happen in the group's own home-cage, thus minimising disruption to their life-cycle.
Apart from the mice, the cage contains a food and drink hopper, bedding and a movable tunnel (enrichment structure), as shown in \cref{FIG_DATA_FRAME}.
For each cage (group of three mice), three to four day continuous recordings are performed when the mice are 3-months, 7-months, 1-year and 18-months old.
During monitoring, the mice are kept on a standard 12-hour light/dark cycle with lights-on at 07:00 and lights-off at 19:00.

\subsubsection{Modalities}
The recordings (\emph{video} and \emph{position}) are split into 30-minute \emph{segments} to be more manageable.
Experiments are thus uniquely identified by the cage-id to which they pertain, the age group at which they are recorded and the segment number.

A single-channel infra-red camera captures video at 25 frames per second from a side-mounted viewpoint in $1280 \times 720$ resolution.
Understandably, the hopper itself is opaque and this impacts the lighting (and ability to resolve objects) in the lower right quadrant.
As regards cage elements, the hopper itself is static, and the mice can feed either from the left or right entry-points.
The water-spout is on the left of the hopper towards the back of the cage from the provided viewpoint.
The bedding itself consists of shavings and is highly dynamic, with the mice occasionally burrowing underneath it.
Similarly, the cardboard tunnel roll can be moved around or chewed and varies in appearance throughout recordings.
This clutter, together with the close confines of the cage, lead to severe occlusion, even between the mice themselves.

With no visual markings, the mice are only identifiable through the implanted \gls{rfid} tag, which is picked up by a $3\times 6$ antenna-array below the cage.
For visualisation purposes (and ease of reference), mice within the same cage are sorted in ascending order by their identifier and denoted Red, Green and Blue.
The antennas are successively scanned in numerical order to test for the presence of a mouse: the baseplate does on average 2.5 full-scans per-second, but this is synchronised to the video frame rate.
The \gls{rfid} pickup itself suffers from occasional dropout, especially when the mice are above the ground (\eg climbing or standing on the tunnel) or in close proximity (\ie during huddling).

\subsubsection{Identifying the Mice}
A key challenge in the use of the home-cage data is the correct tracking and identification of each individual in the group.
This is necessary to relate the behaviour to the individual and also to connect statistics across recordings (including different age-groups).
However, the close confines of the cage and lack of visible markers make this a very challenging problem.
Indeed, standard methods, including the popular DeepLabCut framework of \citet{VL_069} do not work on the kind of data that we use.

Our solution lies in the use of the \gls{tim}, documented in \citet{VL_083}.
We leverage \glspl{bb} output by a neural network mouse detector, which are assigned to the weak location information by solving a custom covering problem.
The assignment is based on a probabilistic weight model of visibility, which considers the probability of occlusion.
The \gls{tim} yields per-frame identified \glspl{bb} for the visible mice and an indication when it is not visible otherwise.

\subsection{\acrshort{imadge}: A dataset for Behaviour Analysis}\label{SS_DATA_IMADGE}
The \gls{imadge} is our curated selection of data with the aim to provide a general dataset for analysing mouse behaviour in group settings.
It includes automatically-generated localisation and behaviour labels for the mice in the cage, and is available at \urlimadge\ for research use.
The dataset also forms the basis for the \acrshort{abode} dataset (\cref{SS_DATA_ABODE}).

\subsubsection{Data Selection}
\gls{imadge} contains recordings of mice from 15 cages from the Adult (1-year) and 10 cages from the Young (3-month) age-groups: nine of the cages exist in both subsets and thus are useful for comparing behaviour dynamics longitudinally.
All mice are male of the \strain strain.
Since this strain of mice is crepuscular (mostly active at dawn/dusk), we provide segments that overlap to \emph{any} extent with the morning (06:00-08:00) and evening (18:00-20:00) periods (at which lights are switched on or off respectively), resulting in generally 2\nicefrac{1}{2} hour recording \emph{runs}.
This is particularly relevant, because changes in the onset/offset of activity around these times can be very good early predictors of \eg neurodegenerative conditions \citep{PHT_021}.
The runs are collected over the three-day recording period, yielding six runs per-cage, equivalent to 90 segments for the Adult and 61 segments for the Young age-groups.

\subsubsection{Data Features}
\gls{imadge} exposes the raw video for each of the segments.
The basic unit of processing for all other features, is the \gls{btp} which is one-second in duration (25 video frames).
This was chosen to balance expressivity of the behaviours (reducing the probability that a \gls{btp} spans multiple behaviours) against imposing an excessive effort in annotation for training behaviour classifiers).

The main modality is the per-mouse behaviour, obtained automatically by our \glsfirst{alm}.
The observability of each mouse in each \gls{btp} is first determined: behaviour classification is then carried out on samples deemed \OBS.
The behaviour is according to one of seven labels: \ALLLBLS{and}.
Behaviours are mutually exclusive within the \gls{btp}, but we retain the full probability score over all labels rather than a single class label.

The \gls{rfid}-based mouse position per-\gls{btp} is summarised in two fields: the mode of the pickups within the \gls{btp} and the absolute number of antenna cross-overs.
The \glspl{bb} for each mouse are generated per-frame using our own \gls{tim} \citep{VL_083}, running on each segment in turn.
The per-\gls{btp} \gls{bb} is obtained by averaging the top-left/bottom-right coordinates throughout the \gls{btp} (for each mouse).

\subsection{\acrshort{abode}: A dataset for Behaviour Classification}\label{SS_DATA_ABODE}
Our analysis pipeline required a mouse behaviour dataset that can be used to train models to automatically classify behaviours of interest, thus allowing us to scale behaviour analysis to larger datasets.
Our answer to this need is the \gls{abode}.
The dataset, available at \urlabode\ consists of video, per-mouse locations in the frame and per-second behaviour labels for each of the mice.

\subsubsection{Data Selection}
For \gls{abode} we used a subset of data from the \gls{imadge} dataset.
We \emph{randomly} selected 200 two-minute snippets from the Adult age-group, with 100 for Training, 40 for Validation and 60 for Testing.
These were selected such that data from a cage appears exclusively in one of the splits (training/validation/test), ensuring a better estimate of generalisation performance.
The data was subsequently annotated by a trained \PHENO (see appendix \cref{APP_ABODE}).

\begin{figure*}
  \centering
  \includegraphics[width=0.9\linewidth]{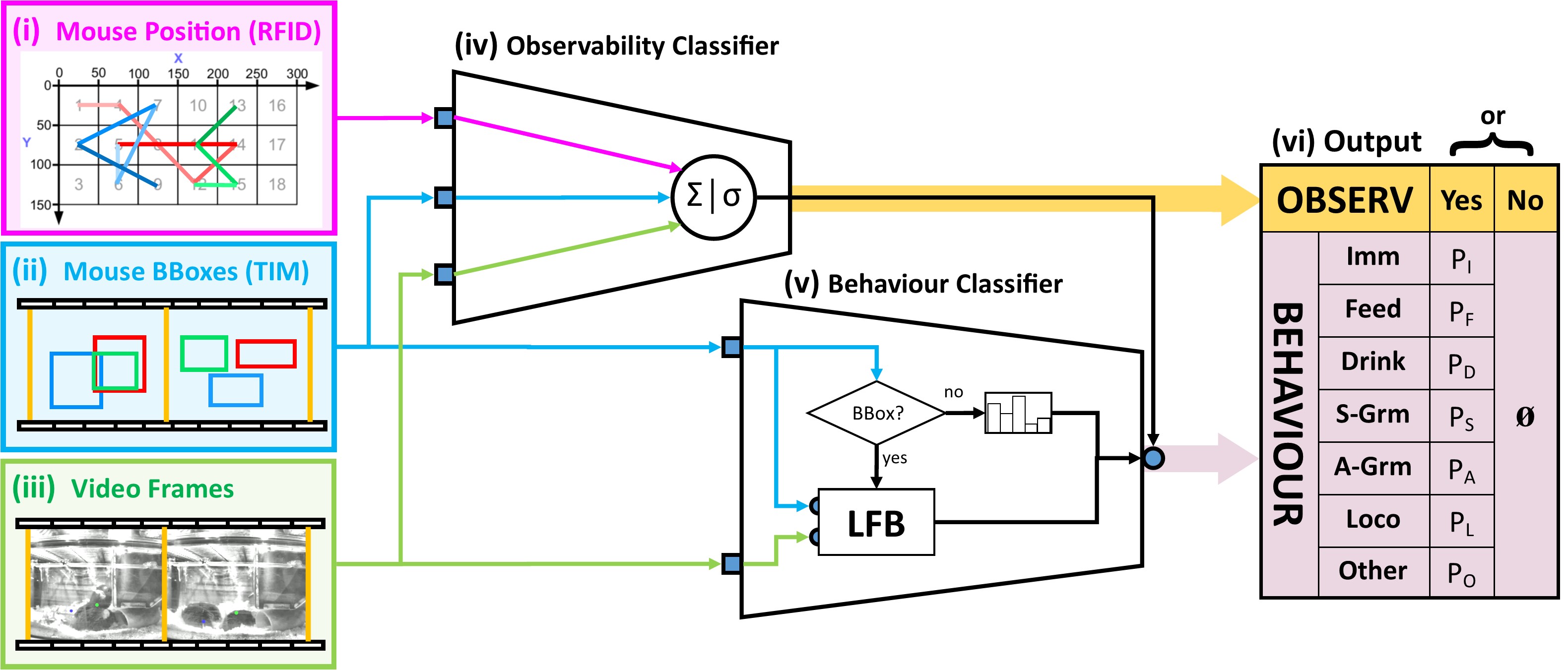}
  	\caption{
  		The \gls{alm} for classifying observability and behaviour \emph{per mouse}.
  		The input signal comes from three modalities: (i) coarse position (\gls{rfid}), (ii) identified \glspl{bb} (using the \gls{tim} as implemented in~\citep{VL_083}) and (iii) video frames.
  		An \glsfirst{oc} (iv) determines whether the mouse is observable and its behaviour can be classified.
  		If this is the case, then the \glsfirst{bc} (v) is activated to generate a probability distribution over behaviours for the mouse.
  		Further architectural details appear in the text.
  	}
  \label{FIG_ALM}
\end{figure*}

\subsubsection{Data Features}
As in \gls{imadge}, \gls{abode} contains the raw video (as two-minute snippets).
We also provide the \emph{per-frame} per-mouse \gls{rfid}-based position reading and \gls{bb} location within the image (generated using \gls{tim}).

The dataset consists of 200 two-minute snippets, split as 110 Training, 30 Validation and 60 in the Test set (see \cref{TAB_ABODE_STATS}).
To simplify our classification and analysis, the behaviour of each mouse is defined at regular \glspl{btp}, and is either \NOOB or one of seven mutually exclusive labels: \ALLLBLS{or}.
We enforce that each \gls{btp} for each mouse is characterised by exactly one behaviour: this implies both exhaustibility and mutual exclusivity of behaviours.
The behaviour of each mouse is annotated by a trained \PHENO according to a more extensive labelling schema which takes into account tentative labellings and unclear annotations: this is documented in the appendix \cref{SS_APP_ANNOTATION}.
Note that unlike some other group-behaviour projects, we focus on individual actions (as above) rather than explicitly positional interactions (\eg Nose-to-Nose, Chasing, \etc) --- this is a conscious decision that is driven by the biological processes under study as informed through years of research experience at the \gls{mrc}.

\section{Methods}\label{S_METHODS}

The main contribution of this work relates to the \gls{gbm} which is described in \cref{SS_METHOD_MODELLING}.
However, obtaining behaviour labels for each of the mice necessitated development of the \gls{alm}, described in \cref{SS_METHOD_ALM}.

\subsection{Classifying Behaviour: the \acrshort{alm}}\label{SS_METHOD_ALM}

Analysing behaviour dynamics in social settings requires knowledge of the individual behaviour throughout the observation period.
Our goal is thus to label the activity of each mouse or flag that it is \NOOB at discrete \glspl{btp} --- in our case every second.
A strong-point of our analysis is the volume of data we have access to: this allows our observations to carry more weight and be more relevant to the biologists as they are drawn from hours (rather than minutes) of observations.
However, this scale of data is also challenging, making manual labelling infeasible.

As already argued in \cref{SS_REL_STPS,SS_REL_BC}, existing setups do not consider the side-view home-cage environment that we deal with.
It was thus necessary to develop our own \glsfirst{alm} (\cref{FIG_ALM}), to automatically determine whether each mouse is observable in the video, and if so, infer a probability distribution over which behaviour it is exhibiting.
Using discrete time-points simplifies the problem by framing it as a purely classification task, and making it easier to model (\cref{SS_METHOD_MODELLING}).
We explicitly use a hierarchical label space (observability v.\ behaviour, see \cref{FIG_ALM}(vi)), since (a) it allows us to break down the problem using an \gls{oc} followed by a \gls{bc} in cascade, and (b) because we prefer to handle \NOOB explicitly as missing data rather than having the \gls{bc} infer unreliable classifications which can in turn bias the modelling.
It is also semantically inconsistent to treat \NOOB as a mutually exclusive label with the rest of the behaviours: specifically, if the mouse is \NOOB, we know it is doing exactly one of the other behaviours (even if we cannot be sure about which).

In the next subsections we describe in turn the \gls{oc} and \gls{bc} sub-modules: note that we postpone detailed training and experimental evidence for the choice of the architectures to our Experiments \cref{S_EXP}.

\subsubsection{Determining Observability} \label{SSS_METHOD_OC}
For the \gls{oc} (iv in \cref{FIG_ALM}) we use as features: the position of the mouse (\gls{rfid}), the fraction of frames (within the \gls{btp}) in which a \gls{bb} for the mouse appears, the average area of such \glspl{bb} and finally, the first 30 \gls{pca} components from the feature-vector obtained by applying the \gls{lfb} model~\citep{VL_082} to the video.
These are fed to a logistic-regression classifier trained using the binary cross-entropy loss \citep[206]{MISC_023} with $l_2$ regularisation, weighted by inverse class frequency (to address class imbalance).
We judiciously choose the operating point (see \cref{SSS_EXP_OC}) to balance the errors the system makes.
Further details regarding the choice and training of the classifier appear in \cref{SSS_EXP_OC}.

\subsubsection{Probability over Behaviours} \label{SSS_METHOD_BC}
The \gls{bc} (v in \cref{FIG_ALM}) operates only on samples deemed \OBS by the \gls{oc}, outputting a probability distribution over the seven behaviour labels (\cref{SS_DATA_ABODE}).
The core component of the \gls{bc} is the \gls{lfb} architecture of \citet{VL_082} which serves as the backbone activity classifier.
For each \gls{btp}, the centre frame and six others on either side at a stride of eight are combined with the first detection of the mouse in the same period and fed to the \gls{lfb} classifier.
The logit outputs of the \gls{lfb} are then calibrated using temperature scaling \citep{NA_031}, yielding a seven-way probability vector.
In instances where there is no detection for the \gls{btp}, a default distribution is output instead.
All components of the \gls{bc} (including choice of backbone architecture) were finetuned on our data as discussed in \cref{SSS_EXP_BEHAVIOUR}.

Although the identification of key-points on a mouse is a sensible way to extract pose information in a clean environment with a top-mounted camera, it is much more problematic in our cluttered home-cage environment with a side-mounted camera.
Indeed, attempts to use the popular DeepLabCut framework \citep{VL_069} failed because of the lack of reproducible key points (see previous work in \citealt*[sec.\ 5.2]{VL_083}).
Hence we predict the behaviour with the \gls{bc} directly from the \gls{rfid} data, \glspl{bb} and frames (as illustrated in \cref{FIG_ALM}), without this intermediate step.

\begin{figure}
  \centering  
  \includegraphics[width=0.5\linewidth]{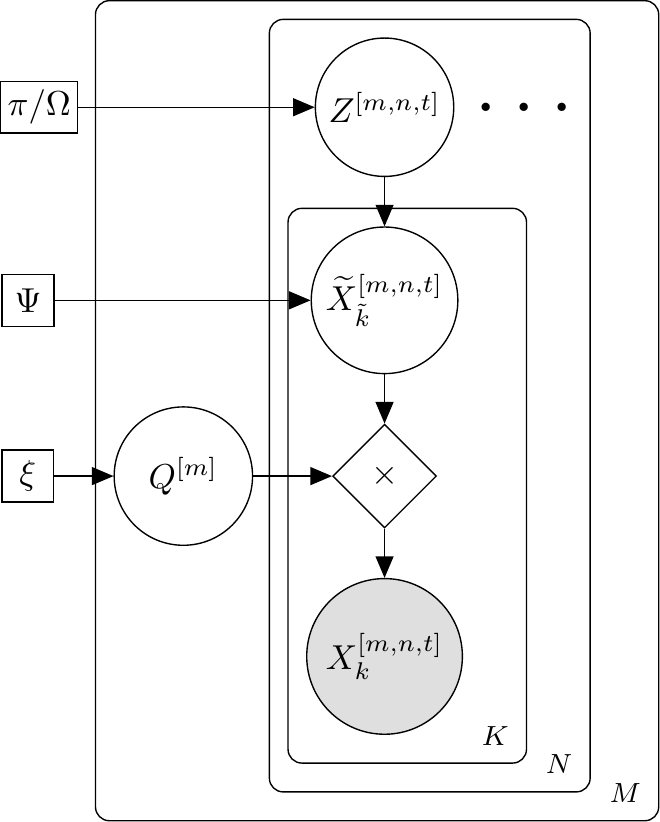}
  \caption{Graphical representation of our \acrshort{gbm}. `$\times$' refers to standard matrix multiplication. To reduce clutter, the model is not shown unrolled in time.}
  \label{FIG_MDL_GLOBAL_GRAPH}
\end{figure}

\subsection{Modelling Behaviour Dynamics}\label{SS_METHOD_MODELLING}
In modelling behaviour, we seek to:
(a) capture the temporal aspect of the individual's behaviour, \textsl{and}
(b) model the interaction and interdependence between cage-mates.
These goals can be met through fitting a \gls{hmm} on a per-cage basis, in which the behaviour of each mouse is represented by factorised categorical emissions contingent on a latent `regime' (which couples them together).
However, this generates a lot of models, making it hard to analyse and compare observations across cages.

To address this, we seek to fit one \glsfirst{gbm} across cages.
The key problem is that the assignment of mouse identities in a cage (denoted as R, G, B) is arbitrary.
As an example, if R represents a dominant mouse in one cage, this role may be taken by \eg mouse G in another cage\footnote{Note that this is just an example for illustrative purposes: the \gls{gbm} makes \emph{no} assumption about any social relations within the cage.}.
Forcing the same emission probabilities across mice avoids this problem, but is too restrictive of the dynamics that can be modelled.
Instead, we introduce a permutation matrix to match the mice in any given cage to the \gls{gbm} as shown in \cref{FIG_MDL_GLOBAL_GRAPH}.
This formulation is broadly applicable to scenarios in which one seeks to uncover shared behaviour dynamics across different entities (\eg in the analysis of sports plays).

As in a \gls{hmm}, there is a latent state $Z$ indexed by cage $m$, recording-run $n$ and time $t$, forming a Markov chain (over $t$), which represents the state of the cage as a whole.
This `regime', is parametrised by $\PZ$ in the first time-point (initial probability) as well as $\TZ$ (transition probabilities), and models dependence both in time as well as between individuals.
We then use $\widetilde{X}$ to denote the behaviour of each mouse: this is a vector of variables, one for each mouse $\tilde{k} \in \lbrace 1, \dots, K\rbrace$, in which the order follows a `canonical' assignment.
Note that each mouse is represented by a complete categorical probability distribution (as output from the \gls{alm}), rather than a hard label, and is conditioned on $Z$ through the emission probabilities $\PX$.
This allows us to propagate uncertainty in the outputs of the \gls{alm} module, with the error implicitly captured through \PX.

\begin{algorithm*}
\caption{Modified \gls{em} for \gls{gbm}. Equations appear in the appendix.}          
\label{ALGO_MDL_GLOBAL_EM}                           
\begin{algorithmic}[1]
\Require {$X$}\Comment{Observations for all cages}
\Require {$\hat{\PQ}, \hat{\PX}, \hat{\PZ}, \hat{\TZ}$} \Comment{Initial Parameter Estimates}
\Repeat
	\ForAll{cages $m \in M$}
		\State{$\hat{q}^{[m]} \gets \argmax_{q' \in Q^{[m]}}P\left(q'|X\right)$} \Comment{\cref{EQ_DERIVE_POST_Q_DEF}}
		\State{Compute $\widetilde{X}^{[m]}$ given $\hat{q}^{[m]}$} \Comment{\cref{EQ_MDL_GLOBAL_XTILDE}}
	\EndFor
	\State{\textbf{E-Step:} Compute Posterior Statistics for $Z$ ($\gamma$, $\eta$)} \Comment{Eqs.\ \eqref{EQ_DERIVE_HMM_GAMMA_RECURSE} and \eqref{EQ_DERIVE_HMM_ETA_RECURSE}}
	\State{\textbf{M-Step:} Update parameters $\hat{\PX}, \hat{\PZ}$ and $\hat{\TZ}$} \Comment{Eqs.\ \eqref{EQ_MOC_M_PI_MLE}, \eqref{EQ_MOC_M_PSI_MLE} and \eqref{EQ_MDL_HMM_TZ_UPDATE}}
	\State{Compute Log-Likelihood using new $\hat{\PQ}, \hat{\PX}, \hat{\PZ}, \hat{\TZ}$} \Comment{\cref{EQ_DERIVE_HMM_OBS_LL}}
\Until{Change in Log-Likelihood $<$ Tolerance}
\State{Re-Optimise Permutation} \Comment{(Lines 2: to 5:)}
\end{algorithmic}
\end{algorithm*}

For each cage $m$, the random variable $Q^{[m]}$ governs which mouse, $k$ (R/G/B) is assigned to which index, $\smash{\tilde{k}}$, in the canonical representation $\smash{\widetilde{X}}$, and is fixed for all samples $n,t$ and behaviours $x$.
The sample space of $Q$ consists of all possible permutation matrices of size $K\times K$ \ie matrices whose entries are 0/1 such that there is only one `on' cell per row/column.
$Q$ can therefore take on one of $K!$ distinct values (permutations).
This permutation matrix setup has been used previously, \eg for problems of data association in multi-target tracking (see, \eg, \citealt*[sec.\ 29.9.3]{murphy-23}), and in static matching problems, see \eg, \citet{mena-etal-20}, \citet{powell-smith-19} and \citet{VL_093}.
In the above cases the focus is on approximate matching due to a combinatorial explosion, but here we are able to use exact inference due to the low dimensionality (in our case $|Q| = 3! = 6$).
This is because the mouse identities have already been established though time in the \gls{tim}, and it is only the permutation of roles between cages that needs to be considered.
The \gls{gbm} is a novel combination of a \gls{hmm} to model the interdependence between cage-mates, and the use of the permutation matrix to handle the mapping between the model's canonical representation $\tilde{X}$ and the observed $X$.

Note that fixing $Q$ and $X$ determines $\smash{\widetilde{X}}$ completely by simple linear algebra.
This allows us to write out the complete data likelihood as:
\begin{align}
P_\Theta\left(\SData\right) & = \prod_{m, n} \Biggl( P_\PQ\left(Q^{[m]}\right) 
P_\PZ\left(Z^{[.,1]}\right)  \prod_{t=2}^{T^n} P_\TZ\left(Z^{[.,t]}|Z^{[.,t-1]}\right) \prod_{t=1}^{T^n}P_\PX\left(X^{[.,t]}|Z^{[.,t]},Q^{[m]}\right) \Biggr) . \label{EQ_MDL_GLOBAL_LIKELIHOOD}
\end{align}

The parameters $\Theta = \left\lbrace \PZ, \TZ, \PQ, \PX\right\rbrace$ of the model are inferred through the \gls{em} algorithm~\citep{AMD_015} as shown in \cref{ALGO_MDL_GLOBAL_EM} and detailed in the appendix.
We seed the parameter set using a model fit to one cage, and subsequently iterate between optimising $Q$ (per-cage) and optimising the remaining parameters using standard EM on the data from all cages.
This procedure is carried out using models initialised by fitting to each cage in turn, and then the final model is selected based on the highest likelihood score (much like with multiple random restarts).
Furthermore, we use the fact that the posterior over $Q$ is highly peaked, to replace the expectation over $Q$ by its maximum (a point estimate), thereby greatly reducing the computational complexity.


\section{Experiments} \label{S_EXP}
We report two sets of experiments.
We begin in \cref{SS_EXP_ALM} by describing the optimisation of the various modules that make up the \gls{alm}, and subsequently, describe the analysis of the group behaviour in \cref{SS_EXP_GBA}.
The code to produce these results is available at \CODE.

\subsection{Fine-tuning the \acrshort{alm}}\label{SS_EXP_ALM}
The \gls{alm} was fit and evaluated on the \gls{abode} dataset.

\subsubsection{Metrics}
For both the observability and behaviour components of the \gls{alm} we report accuracy and \FS score \citep*[see \eg][Sec.~5.7.2.3]{MISC_013}.
We use the macro-averaged \FS to better account for the class imbalance.
This is particularly severe for the observability classification, in which only about 7\% of samples are \NOOB, but it is paramount to flag these correctly.
Recall that the \OBS samples will be used to infer behaviour (\cref{SSS_METHOD_BC}) which is in turn used to characterise the dynamics of the mice (\cref{SS_METHOD_MODELLING}).
Hence, it is more detrimental to give \gls{fp} outputs, which results in \UNREL behaviour classifications (\ie when the sample is \NOOB but the \gls{oc} deems it to be \OBS, which can throw the statistics awry) than missing some \OBS periods through \glspl{fn} (which, though \WASTE of data, can generally be smoothed out by the temporal model).
This construct is formalised in \cref{TAB_OBS_METRICS}, where we use the terms \UNREL and \WASTE as they better illustrate the repercussions of the errors.
In our evaluation, we report the number of \UNREL and \WASTE samples to take this imbalance into account.
For the \gls{bc}, we also report the normalised (per-sample) log-likelihood score, \smash{\nll}, given that we use it as a probabilistic classifier.

\begin{table*}[!h]
\caption{Definition of classification outcomes for the Observability problem.
GT refers to the ground-truth (annotated) and the standard machine learning terms --- \gls{tp}, \gls{fp}, \gls{tn}, \gls{fn} --- are in square brackets.}
\label{TAB_OBS_METRICS}
\begin{center}
\begin{tabular}{c|ccc}
	\multicolumn{1}{c}{}    		 &                        & \multicolumn{2}{c}{\textsc{Predicted}} \\
																\cmidrule{3-4}
	\multicolumn{1}{c}{}    		 & 					      & \textbif{Obs.}    & \textbif{N/Obs.} \\
	\multirow{2}{*}{\textsc{GT}} & \textbif{Obs.}     & True \OBS [\acrshort{tp}]  & \WASTE [\acrshort{fn}] \\
					    			 & \textbif{N/Obs.} & \UNREL [\acrshort{fp}]     & True \NOOB [\acrshort{tn}]\\
\end{tabular}
\end{center}
\end{table*}

\subsubsection{Observability} \label{SSS_EXP_OC}
\begin{figure}
\centering
\includegraphics[width=0.6\linewidth]{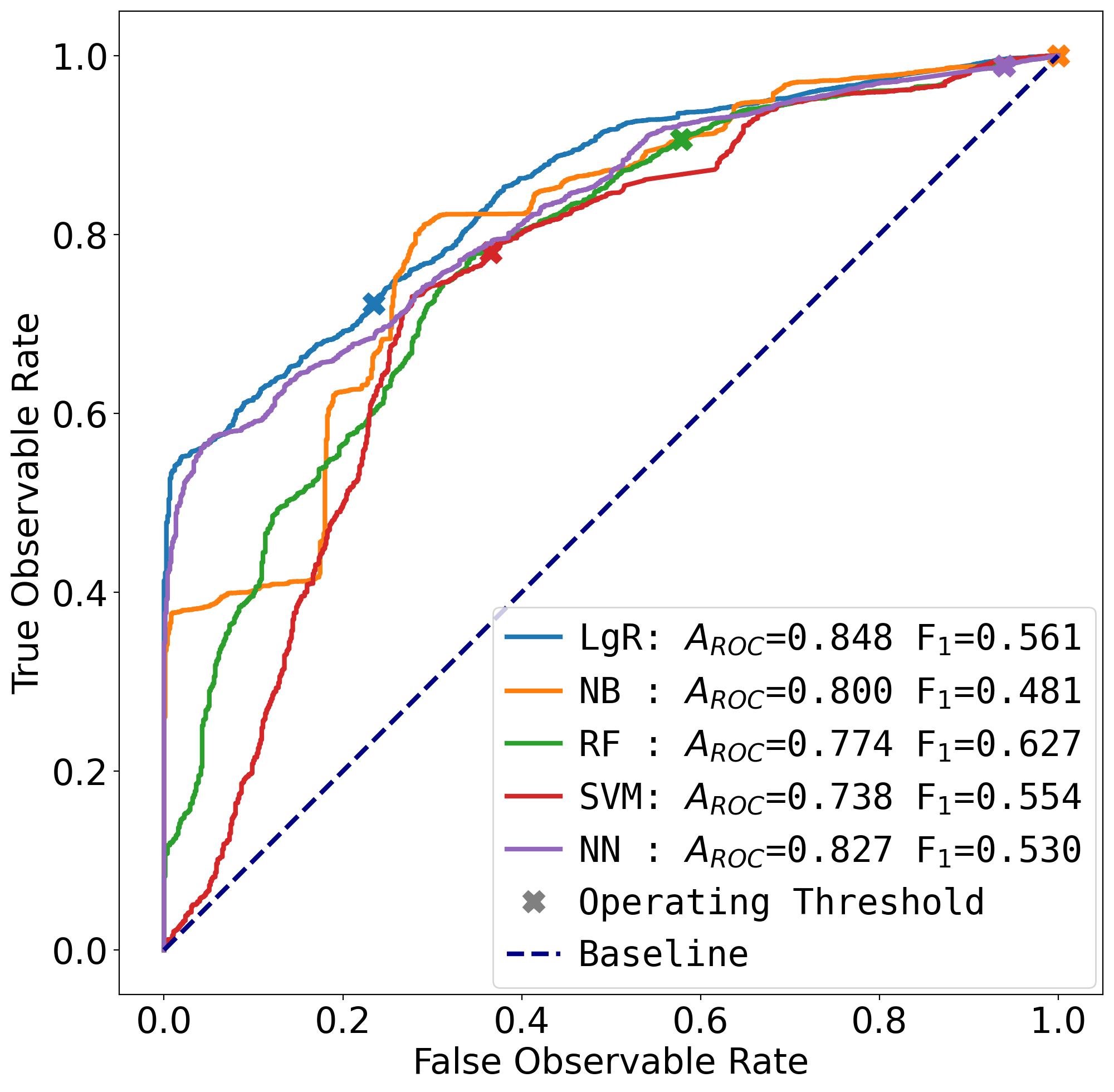}
\caption{ROC curves for various architectures of the \gls{oc} evaluated on the Validation split. 
Each coloured line shows the \gls{tp} rate against the \gls{fp} rate for various operating thresholds: the `default' 0.5 threshold in each case is marked with a cross `$\times$'.
The baseline (worst-case) model is shown as a dotted line.}
\label{FIG_ROC_OC}
\end{figure}

The challenge in classifying observability was to handle the severe class imbalance, which implied judicious feature selection and classifier tuning.
Although the observability sample count is high within \gls{abode}, the skewed nature (with only 7\% \NOOB) is prone to overfitting.
Features were selected based on their correlation with the observability flag, and narrowed down to the subset already listed (\cref{SSS_METHOD_OC}).
As for classifiers, we explored \gls{lgr}, \gls{nb}, Random Forests (RF), Support-Vector Machines (SVM) and feed-forward Neural Networks (NN).
Visualising the ROC curves \citep[see \eg][Sec.\ 5.7.2.1]{MISC_013} (\cref{FIG_ROC_OC}), brings out two clear candidate models.
Note how at most operating points, the \gls{lgr} model is the best classifier, except for some ranges where \gls{nb} is better (higher).
These were subsequently compared in terms of the number of \UNREL and \WASTE samples at two thresholds: one is at the point at which the number of \WASTE samples is on par with the true number of \NOOB in the data (\ie 8\%), and the other at which the number of predicted \NOOB equals the statistic in the ground-truth data.
These appear in \cref{STAB_EXP_OC}: the \gls{lgr} outperforms the \gls{nb} in almost all cases, and hence we chose the \gls{lgr} classifier operating at the \WASTE = 8\% point.

\begin{table}
\centering
\setlength\tabcolsep{1.2mm}
\caption{Comparison of \gls{lgr} and \gls{nb} as observability classifiers (on the validation set) at different operating points. The best performing score in each category appears in bold. Note that for context, there are 10,124 samples, of which 750 are \NOOB.}
\label{STAB_EXP_OC}
\begin{tabular}{lcccc}
	\toprule[1.5pt]
	{} 		  & \multicolumn{2}{c}{\WASTE = 8\%}    & \multicolumn{2}{c}{Equal \NOOB} \\
         \cmidrule(lr){2-3}                 \cmidrule(lr){4-5}
	{} 		  & \UNREL$\downarrow$ &  \WASTE$\downarrow$ & \UNREL$\downarrow$ &  \WASTE$\downarrow$ \\
	\midrule[1pt]
	\gls{lgr} & \textbf{381} & \textbf{742}  &          499 & \textbf{488} \\
	\gls{nb}  &       470    & 		750      & \textbf{491} &         491  \\
	\bottomrule[1.5pt]
\end{tabular}
\end{table}

\subsubsection{Behaviour}\label{SSS_EXP_BEHAVIOUR}
We explored two architectures as backbones (feature extractors) for the \gls{bc}, the \gls{stlt} of \cite{VL_085} and the \gls{lfb} of \cite{VL_082}, on the basis of them being most applicable to the spatio-temporal action-localisation problem \citep{DS_024}.
In both cases, we:
(a) used pre-trained models and fine-tuned them on our data,
(b) standardised the pixel intensity (over all the images) to unit mean and one standard deviation (fit on samples from the tuning split), \textsl{and}
(c) investigated lighting enhancement techniques \citep{VL_091}, although this did not improve results in any of our experiments.
We provide an overview of the training below, but the interested reader is directed to \citet{MISC_060} for further details.

The \gls{stlt} uses the relative geometry of \glspl{bb} in a scene (layout branch), as well as the video frames (visual branch) to classify behaviour \citep{VL_085}.
The visual branch extracts per-frame features using a ResNet-50,
The layout branch is a transformer architecture which considers the temporal and spatial arrangement of detections of the subject animal and contextual objects --- the other cage-mates and the hopper.
In order to encode invariance to the absolute mouse identities, the assignment of the cage-mates to the slots `cagemate1' and `cagemate2' was randomly permuted during training.
The signal from each branch is fused at the last stages of the classifier.
The base architecture was extended to use information from outwith the \gls{btp}, drawing on temporal context from surrounding time-points.
We ran experiments using the layout-branch only and the combined layout/visual branches, each utilising different number of frames and strides.
We also experimented with focusing the visual field on the detected mouse alone.
Training was done using Adam \citep{MTH_003}, with random-crop and colour-jitter augmentations, and we explored various batch-sizes and learning rates: validation-set evaluation during training directed us to pick the dual-branch model with 25 frames (12 on either side of the centre frame) at a stride of 2 as the best performer.

\begin{table}
\centering
\setlength\tabcolsep{1.2mm}
\caption{Evaluation of the baseline (prior-probability), \gls{stlt} and \gls{lfb} models on the Training and Validation sets in terms of Accuracy, macro-\FS and normalised log-likelihood (\nll). The best performing score in each category (on the validation set) appears in bold.}
\label{TAB_BEH_BEHAVIOUR_MODEL_COMPARE}
\begin{tabular}{lrrrrrr}
	\toprule[1.5pt]
	{} & \multicolumn{3}{c}{Train} & \multicolumn{3}{c}{Validate} \\
     	 \cmidrule(lr){2-4}          \cmidrule(lr){5-7}
	{} &  Acc.\ $\uparrow$ &   \FS$\uparrow$ &  $\nll\uparrow$ &     Acc.\ $\uparrow$ &   \FS$\uparrow$ &  $\nll\uparrow$ \\
	\midrule
	baseline &  0.47 & 0.08 & -1.47 &     0.51 & 0.08 & -1.45 \\
	\gls{stlt}  &  0.77 & 0.45 & -0.70 &     0.73 & 0.36 & \textbf{-1.04} \\
	\gls{lfb}   &  0.96 & 0.93 & -0.11 &     \textbf{0.74} & \textbf{0.61} & -2.27 \\
	\bottomrule[1.5pt]
\end{tabular}
\end{table}

The \gls{lfb} \citep{VL_082}, on the other hand, is a dedicated spatio-temporal action localisation architecture which combines \gls{btp}-specific features with a long-term feature-bank extracted from the entire video, and joined together through an attention mechanism before being passed to a linear classification layer.
Each of the two branches (feature-extractors) uses a FastRCNN network with a ResNet 50 backbone.
We used the pre-trained feature-bank generator and fine-tuned the short-term branch, attention mechanism and classification layer end-to-end on our data.
Training was done using \gls{sgd} with a fixed-step learning scheduler: we used batches of 16 samples and trained for 50 epochs, varying learning rates, warm-up periods and crucially the frame sampling procedure, again in terms of total number of frames and the stride.
We explored two augmentation procedures (as suggested by \citealt{VL_082}): 
(a) random rescaling and cropping, \textsl{and} 
(b) colour jitter (based only on brightness and contrast).
Again, validation-set scores allowed us to choose the 11-frame model with a stride of 8 and with resize-and-crop and colour-jitter augmentations (at the best performing epoch) as our \gls{lfb} contender.

To choose our backbone architecture, the best performer in each case was evaluated in terms of Accuracy, \FS and log-likelihood on both the Training and Validation set in \cref{TAB_BEH_BEHAVIOUR_MODEL_COMPARE}.
Given the validation set scores, the \gls{lfb} model with an \FS of 0.61 (compared to 0.36 for the \gls{stlt}), was chosen as the \gls{bc}.
This was achieved despite the coarser layout information available to the \gls{lfb} and the changes to the \gls{stlt} architecture: we hypothesize that this is due to the ability of the \gls{lfb} to draw on longer-term context from the whole video (as opposed to the few seconds available for the \gls{stlt}).

The \gls{lfb} (and even the \gls{stlt}) model can only operate on samples for which there is a \gls{bb} for the mouse.
We need to contend however with instances in which the mouse is not identified by the \gls{tim}, but the \gls{oc} reports that it should be \OBS.
In this case, we fit a fixed categorical distribution to samples which exhibited this `error' in the training data (\ie \OBS but no \gls{bb}).

\subsubsection{End to End performance}
\begin{figure}[!h]
\centering
\includegraphics[width=0.6\linewidth]{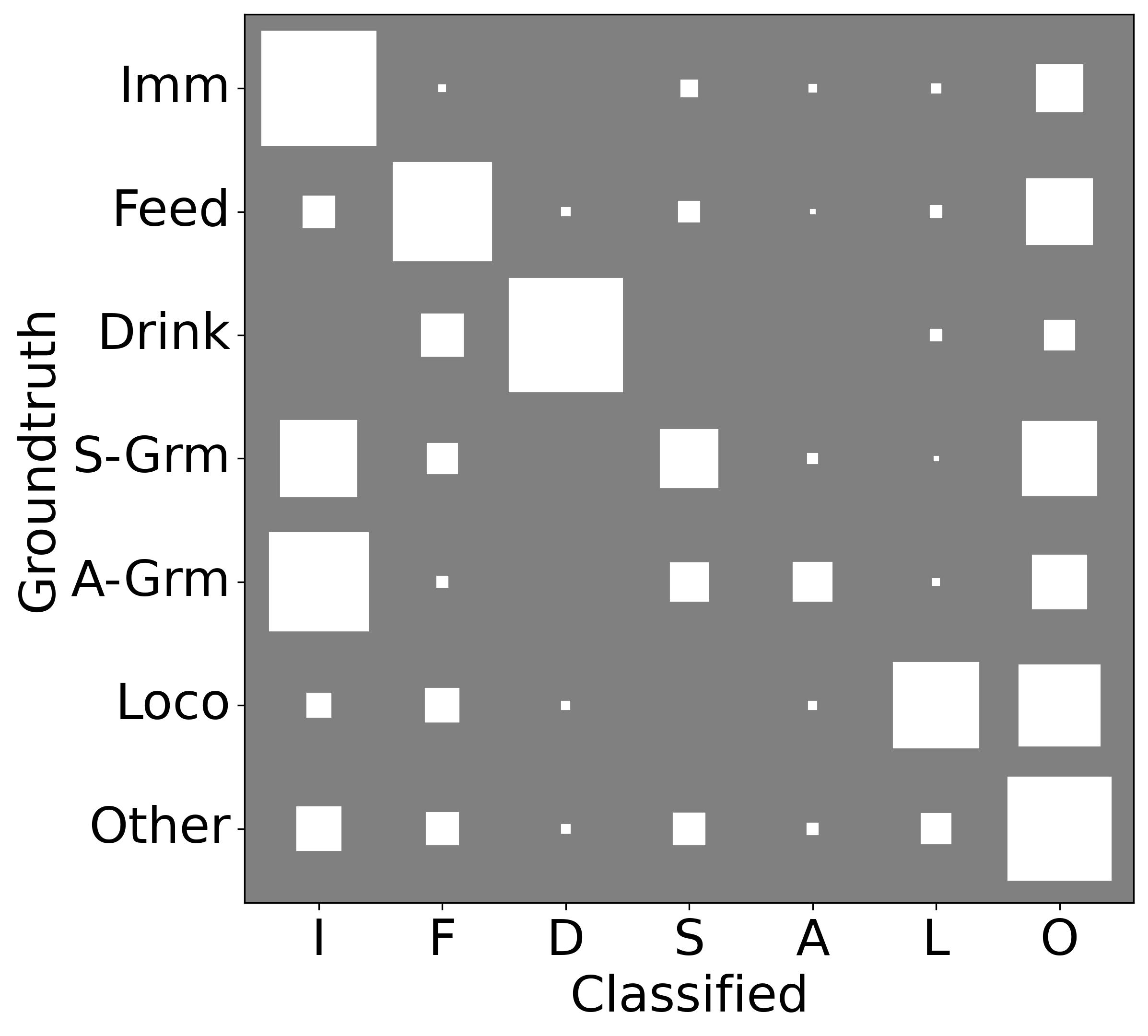}
\caption{Behaviour confusion matrix of the End-to-End model as a Hinton plot. The area of each square represents the numerical value, and each row is normalized to sum to 1.}
\label{FIG_CONFUSION}
\end{figure}

In \Cref{TAB_END2END} we show the performance of the \gls{alm} on the held-out test-set.
Since there is no other system that works on our data to compare against, we report the performance of a baseline classifier which returns the prior probabilities.
In terms of observability, the \gls{alm} achieves slightly less accuracy but a much higher \FS score, as it seeks to balance the types of errors (cutting the \UNREL by 34\%).
In terms of behaviour, when considering only \OBS classifications, the system achieves 68\% accuracy and 0.54 \FS despite the high class imbalance.
The main culprits for the low score are the grooming behaviours, which as shown in \cref{FIG_CONFUSION}, are often confused for \IMM.
Within the supplementary material, we provide a demo video -- Online Resource 1 -- showing the output from the \gls{alm} for a sample 1:00 clip.
In the clip, the mice exhibit a range of behaviours, including \FEED, \LOCO, \SGRM, and \IMM.

\begin{table}
\centering
\setlength\tabcolsep{1.1mm}
\caption{Test performance of the \gls{alm} and baseline model, in terms of observability and behaviour. The best performing score in each category appears in bold. Within the former, U and W refer to the counts of \UNREL and \WASTE respectively: the dataset contains 20,581 samples.}
\label{TAB_END2END}
\begin{tabular}{lrrrrrrr}
	\toprule[1.5pt]
	{}    & \multicolumn{4}{c}{Observability} & \multicolumn{3}{c}{Behaviour}\\
		    \cmidrule(lr){2-5}                  \cmidrule(lr){6-8}
	{}    & Acc.\ $\uparrow$ &  \FS$\uparrow$ & U$\downarrow$ & W$\downarrow$ & Acc.\ $\uparrow$ & \FS$\uparrow$ & $\nll\uparrow$ \\
	\midrule[1pt]
	baseline     & \textbf{0.93} & 0.48 &  1506 	    &  \textbf{0} &      0.48 & 0.09 & -1.47 \\
	\gls{alm} & 0.88 & \textbf{0.61} & \textbf{996} & 1558 	    & \textbf{0.68} & \textbf{0.54} & \textbf{-1.06}\\
	\bottomrule[1.5pt]
\end{tabular}
\end{table}

\subsection{Group Behaviour Analysis}\label{SS_EXP_GBA}
The \gls{imadge} dataset is used for our behaviour analysis, focusing on the adult demographic and comparing with the young one later.

\subsubsection{Overall Statistics}
It is instructive to look at the overall statistics of behaviours.
In \cref{TAB_IMADGE_STATS} we report the percentage of time mice exhibit a particular behaviour, averaged over cages and split by age-group.
The \IMM behaviour clearly stands out as the most prevalent, but there is a marked increase as the mice get older (from 30\% to 45\%) --- this is balanced by a decrease in \OTR, with most other behaviours exhibiting much the same statistics.

\begin{table}
\centering
\setlength\tabcolsep{1.5mm}
\caption{Distribution of Behaviours across cages per age-group.}
\label{TAB_IMADGE_STATS}
\begin{tabular}{lrrrr}
\toprule[1.5pt]
{} & \multicolumn{2}{c}{Young} & \multicolumn{2}{c}{Adult} \\
\cmidrule(lr){2-3} \cmidrule(lr){4-5}
{} & Mean (\%) & Std (\%) & Mean (\%) & Std (\%) \\
\midrule
Imm   &     30.4 &    10.8 &     44.9 &     9.5 \\
Feed  &      7.1 &     1.6 &      8.1 &     1.4 \\
Drink &      0.6 &     0.2 &      0.7 &     0.4 \\
S-Grm &      4.5 &     3.0 &      4.7 &     3.1 \\
A-Grm &      0.6 &     0.5 &      1.1 &     1.0 \\
Loco  &      4.1 &     1.2 &      2.1 &     0.8 \\
Other &     38.5 &    10.2 &     25.9 &     4.1 \\
\bottomrule[1.5pt]
\end{tabular}
\end{table}

\subsubsection{Metrics}
Evaluating an unsupervised model like the \gls{gbm} is not straightforward, since there is no objective ground-truth for the latent states.
Instead, we compare models using the normalised log-likelihood \nll.
When reporting relative changes in \smash{\nll}, we use a baseline model to set an artificial zero (otherwise the log-likelihood is not bounded from below).
Let $\smash{\nll_{BL}}$ represent the normalised log-likelihood of a baseline model: the independent distribution per mouse per frame. 
Subsequently, we use $\smash{\nll_{\SParams}}$ for the likelihood under the global model (parametrised by $\SParams$) and $\smash{\nll_{\SParams}^*}$ for the likelihood under the per-cage model ($\SParams^*$).
We can then define the \gls{rdl} between the two models parametrised as:
\begin{equation}
\text{\gls{rdl}}\left(\SParams;\SParams^*\right) = \frac{\nll_{\SParams} - \nll_{\SParams^*}}{\nll_{\SParams} - \nll_{BL}} \times 100 \% \quad . \label{EQ_MDL_RDL}
\end{equation}

In evaluating on held-out data, we have a further complexity due to the temporal nature of the process.
Specifically, each sample cannot be considered independent with respect to its neighbours.
Instead, we treat each \emph{run} (2\nicefrac{1}{2} hours) as a single fold, and evaluate models using leave-one-out cross-validation: \ie we train on five of the folds and evaluate on the held-out in turn.

\subsubsection{Size of $Z$}
\begin{figure}
  \centering
  \includegraphics[width=0.9\linewidth]{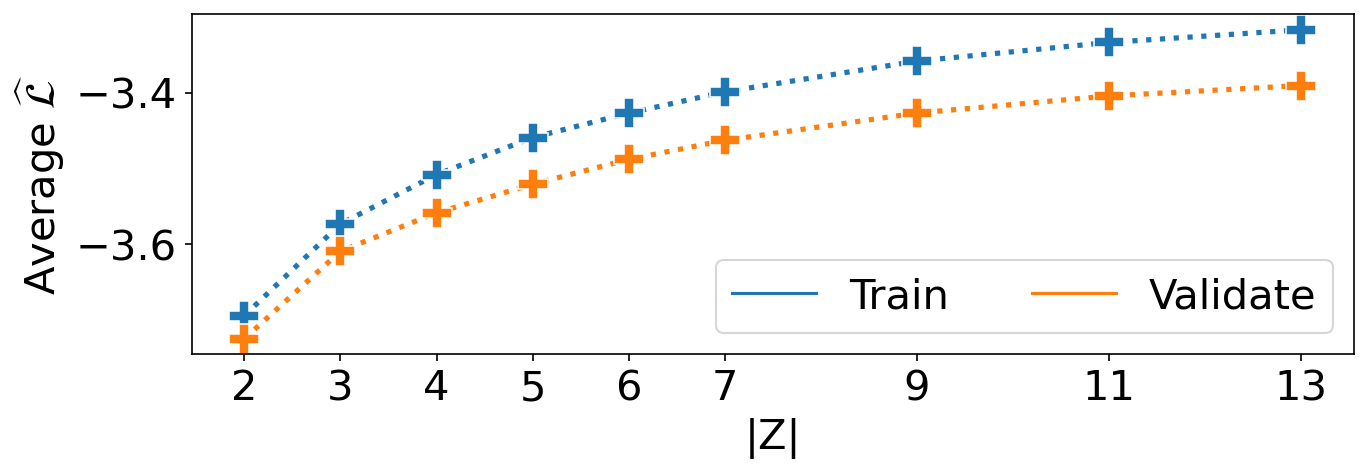}
  	\caption{Normalised log-likelihood (\nll) of the \gls{gbm} for various dimensionalities of the latent state over all cages.}
  \label{SFIG_EXP_GBM_LL_Z}
\end{figure}

The number of latent states $|Z|$ in the \gls{gbm} governs the expressivity of the model: too small and it is unable to capture all the dynamics, but too large and it becomes harder to interpret.
To this end, we fit a per-cage model (\ie without the $Q$ construct) to the adult mice data for varying $|Z| \in \left\lbrace 2, \ldots, 13 \right\rbrace$, and computed \smash{\nll}\ on held out data (using the aforementioned cross-validation approach).
As shown in \cref{SFIG_EXP_GBM_LL_Z}, the likelihood increased gradually, but slowed down beyond $|Z|=7$: we thus use $|Z|=7$ in our analysis.

\subsubsection{Peaked Posterior over $Q$}
\begin{figure}
\centering
\includegraphics[width=0.55\linewidth]{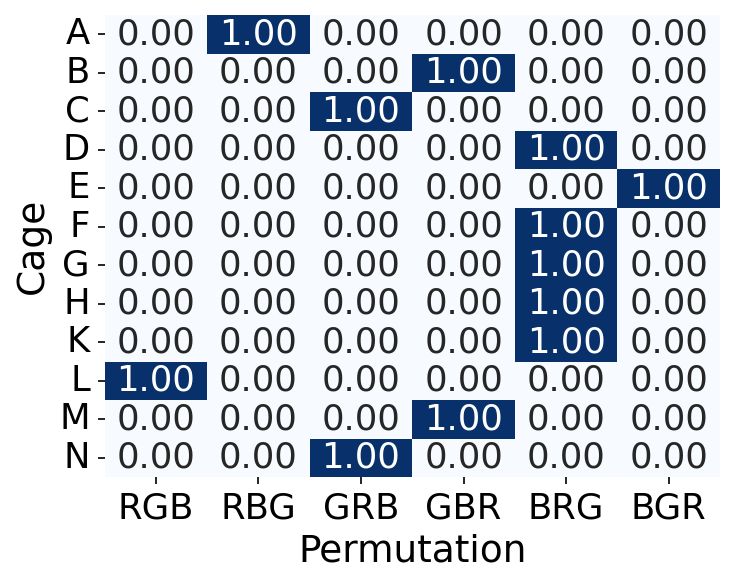}
\caption{Posterior probability of the \gls{gbm} over $Q$ for all cages ($|Z|=7$, model trained on cage L).}
\label{SFIG_EXP_GBM_PEAKEDNESS}
\end{figure}

Our \cref{ALGO_MDL_GLOBAL_EM} assumes that the posterior over $Q$ is sufficiently peaked.
To verify this, we computed the posterior for all permutations over all cages given each per-cage model.
To two decimal places, the posterior is deterministic as shown in \cref{SFIG_EXP_GBM_PEAKEDNESS} for the cages in the Adult demographic using the model trained on cage $L$.
The Young demographic exhibited the same phenomenon.

\subsubsection{Quality of Fit}
\label{SS_QUALITY_FIT}
\begin{figure*}
\centering
\includegraphics[width=0.9\linewidth]{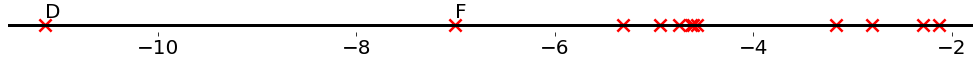}
\caption{\glspl{rdl} from \cref{TAB_EXP_RDL}, printed on the number line: the lowest scoring cages are marked.}
\label{FIG_RDL_NUMBERS}
\end{figure*}

During training of per-cage models, we noticed extremely similar dynamics between the cages (see \citealt*{MISC_060}).
This is not unexpected given that the mice are of the same strain and sex, and recorded at the same age-point.
Nonetheless, we wished to investigate the penalty paid by using a global rather than per-cage model.
To this end, in \cref{TAB_EXP_RDL}, we report the \smash{\nll} and \gls{rdl} (\cref{EQ_MDL_RDL}) of the \gls{gbm} model evaluated on data from \emph{each cage in turn}.
Although the per-cage model is understandably better than the \gls{gbm} on its own data, the average drop is just 4.8\%, which is a reasonable penalty to pay in exchange for a global model.
Plotting the same values on the number-line (\cref{FIG_RDL_NUMBERS}) shows two cages, D and F, that stand out from the rest due to a relatively higher drop.
This led us to further investigate the two cages as potential outliers in our analysis, see \cref{SSS_LATENT_ANALYSIS}.

\begin{table*}
\centering
\setlength\tabcolsep{1.5mm}
\caption{
Evaluation of the \gls{gbm} model ($|Z|=7$) on data from \emph{each} cage (columns) in terms of the Normalised log-likelihood (\nll) and \gls{rdl}.
}
\label{TAB_EXP_RDL}
\begin{tabular}{lcccccccccccc}
\toprule[1.5pt]
 & A & B & C & D & E & F & G & H & K & L & M & N \\
\midrule
$\nll_{\text{\gls{gbm}}}$ & -1.10 & -1.17 & -1.16 & -1.43 & -1.25 & -1.36 & -1.36 & -1.20 & -1.13 & -1.22 & -1.13 & -1.29 \\
\gls{rdl} & -5.32 & -4.56 & -2.29 & -11.14 & -4.94 & -7.00 & -4.63 & -3.17 & -2.81 & -2.13 & -4.61 & -4.75 \\
\bottomrule[1.5pt]
\end{tabular}
\end{table*}

\subsubsection{Latent Space Analysis}
\label{SSS_LATENT_ANALYSIS}
\begin{figure*}
  \centering
  \includegraphics[width=\linewidth]{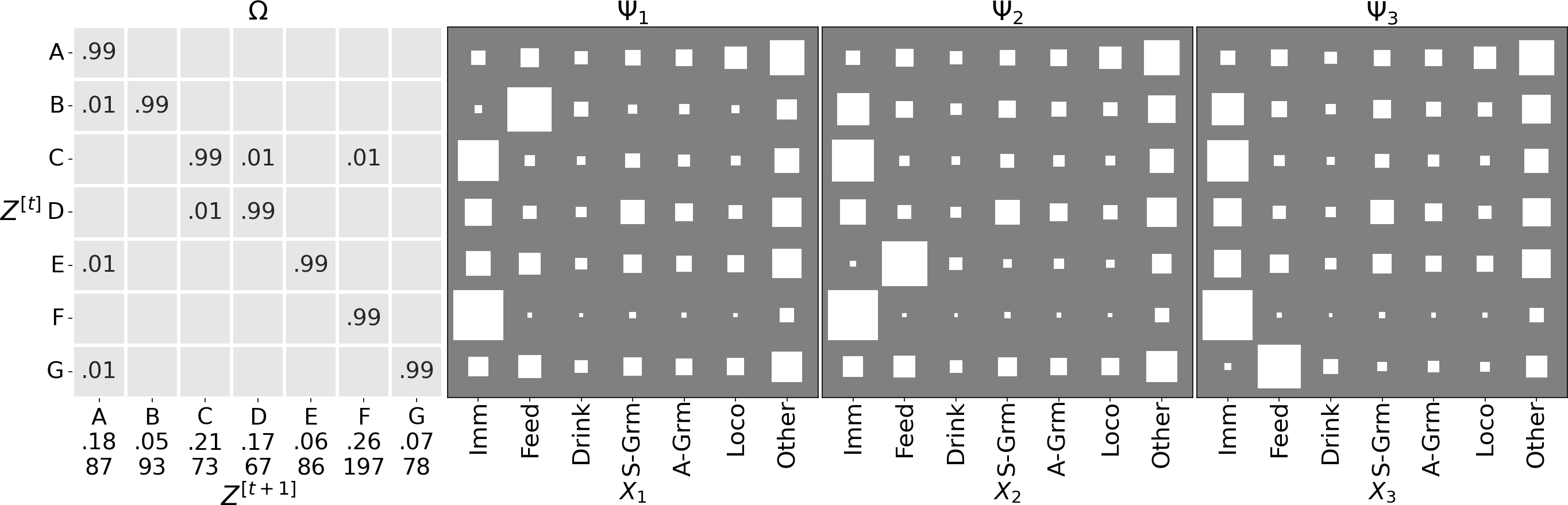}
  \caption{Parameters for the \gls{gbm} with $|Z|=7$ trained on Adult mice.
  	For \TZ\ (leftmost panel) we show the transition probabilities: underneath the $Z^{[t+1]}$ labels, we also report the steady-state probabilities (first row) and the expected dwell times (in \glspl{btp}, second row).
  	The other three panels show the emission probabilities $\PX_k$ for each mouse as Hinton plots.
 	We omit zeros before the decimal point and suppress values close to 0 (at the chosen precision).
 	}
  	\label{FIG_MDL_ADULT_PARAMS}
\end{figure*}

\begin{figure*}
	\centering
	\includegraphics[width=\textwidth]{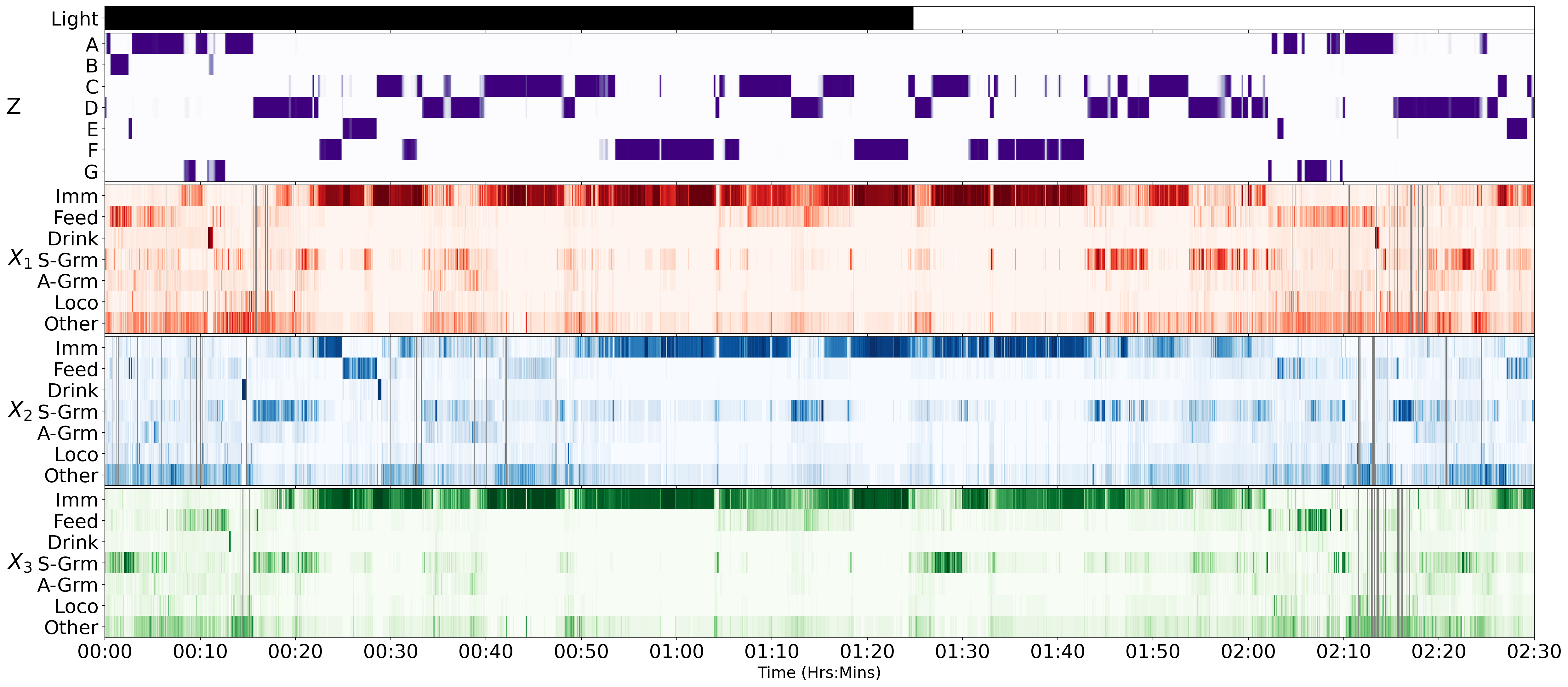}
	\caption{
 	Ethogram for the regime (\gls{gbm} $|Z|=7$) and individual behaviour probabilities for a run from cage B.
 	In all but the light status, darker colours signify higher probability: the hue is purple for $Z$ and matches the assignment of mice to variables $X_k$ otherwise.
 	The light-status is indicated by white for lights-on and black for lights-off.
 	Missing data is indicated by grey bars.
	}
	\label{FIG_MDL_GLOBAL_TEMPORAL_Z7}
\end{figure*}

\Cref{FIG_MDL_ADULT_PARAMS} shows the parameters of the trained \gls{gbm}.
Most regimes have long dwell times, as indicated by the values close to 1 on the diagonal of $\TZ$.
For the emission matrices $\PX_1, \ldots, \PX_3$, note that regime F captures the \IMM behaviour for all mice, and is the most prevalent (0.26 steady state probability) --- it also provides evidence for the anecdotal phenomenon that mice tend to huddle together to sleep.
The purity of this regime indicates that the mice often are \IMM at the same time, re-enforcing the biological knowledge that they tend to huddle together for sleeping, but it is interesting that this was picked up by the model without any apriori bias.
This is further evidenced in the ethogram visualisation in \cref{FIG_MDL_GLOBAL_TEMPORAL_Z7}, which also points out regime C as indicating when any two mice are \IMM, and D for any two mice exhibiting \SGRM.
Similarly, regime A is most closely associated with the \OTR label, although it is less pure.

A point of interest are the regimes associated with the \FEED behaviour, that are different across mice --- B, E and G for mice 1, 2 and 3 respectively.
This is surprising given that more than one mouse can feed at a time (the design of the hopper is such that there is no need for competition for feeding resources).
This is significant, given that it is a global phenomenon, as it could be indicative of a pecking order in the cage.
Another aspect that emerges is the co-occurrence of \SGRM with \IMM or \OTR behaviours: note how in regime (D) (which has the highest probability of \SGRM) these are the most prevalent.

\begin{figure*}
  \centering
  	\begin{subfigure}{\linewidth}
      \includegraphics[width=\linewidth]{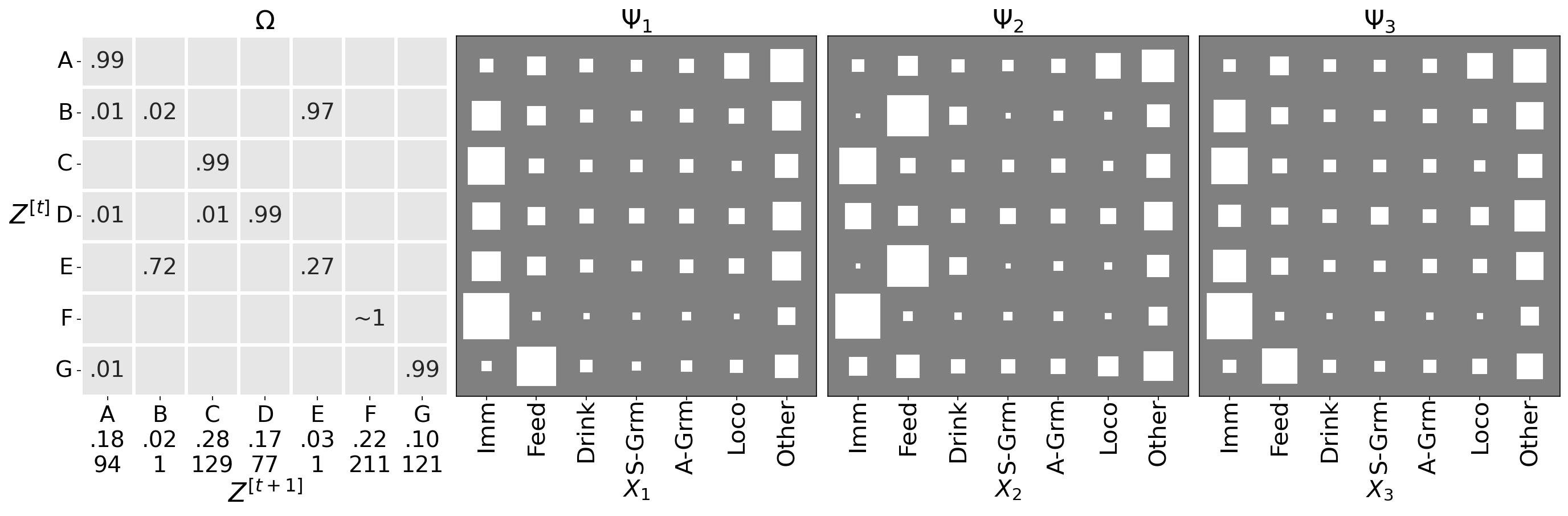}
  	  \vspace{-2em}
    	\caption{Cage D}
  	\end{subfigure}
  	\begin{subfigure}{\linewidth}
      \includegraphics[width=\linewidth]{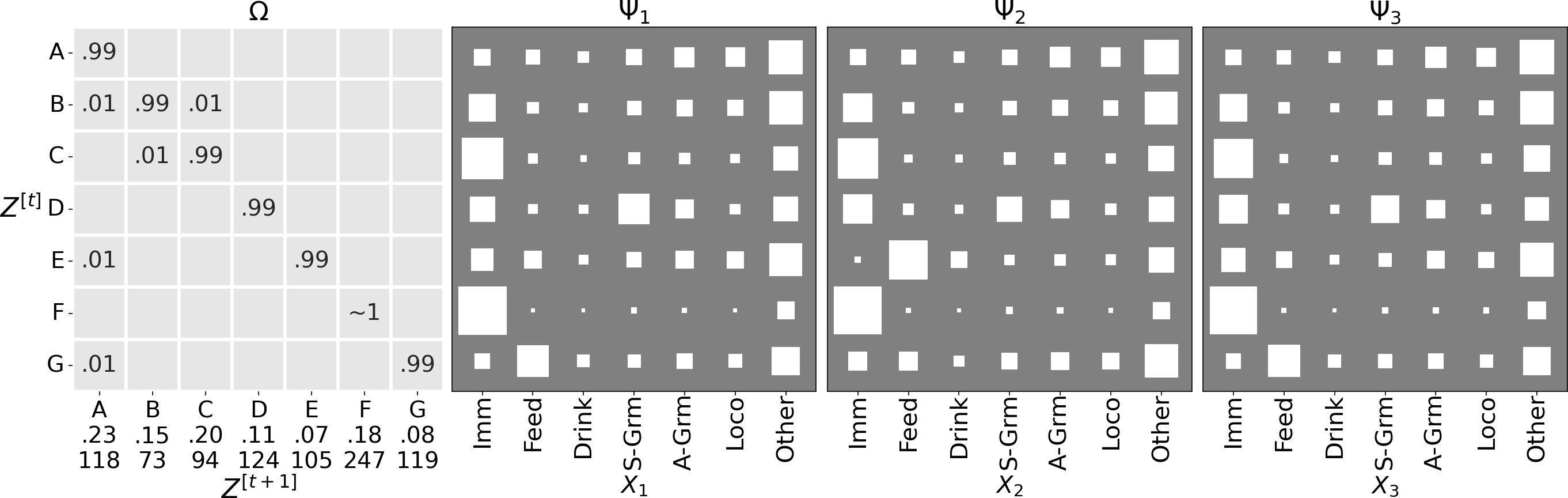}
  	  \vspace{-2em}
    	\caption{Cage F}
  	\end{subfigure}
  	\begin{subfigure}{\linewidth}
      \includegraphics[width=\linewidth]{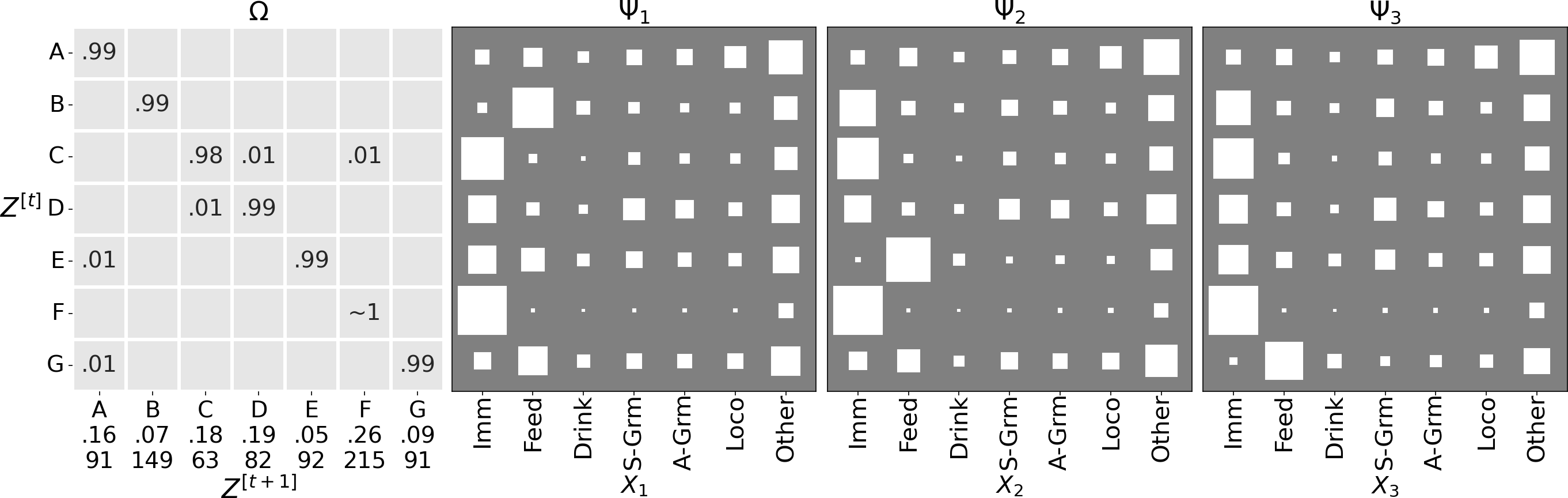}
  	  \vspace{-2em}
    	\caption{Cage L}
  	\end{subfigure}
  \caption{Parameters for the per-cage models ($|Z|=7$) for cages D, F and L.
  	The order of the latent states is permuted to maximise the similarity with the global model (using the Hungarian algorithm) for easier comparison.
  	The plot follows the arrangement in \cref{FIG_MDL_ADULT_PARAMS}.}
  	\label{FIG_MDL_PARAMS_OUTLIERS}
\end{figure*}

Our Quality-of-Fit analysis (\cref{SS_QUALITY_FIT}) highlighted cages D and F as exhibiting deviations in the behaviour dynamics, compared to the rest of the cages.
To investigate these further, we plotted the parameters of the per-cage model for these two cases in \cref{FIG_MDL_PARAMS_OUTLIERS}, and compare them against an `inlier' cage L.
Note that both the latent states and the mouse identities are only identifiable subject to a permutation (this was indeed the need for the \gls{gbm} construct).
To make comparison easier, we optimised the permutations of both the latent states and the ordering of mice that (on a per-cage basis) maximise the agreement with the global model.
The emission dynamics ($\Psi$) for the `outliers' are markedly different from the global model.
Note for example how the feeding dynamics for cages D and F do not exhibit the same pattern as in the \gls{gbm} and cage L: in both these cases the same feeding regime G is shared by mice 1 and 3.
In the case of cage D, there is also evidence that a 6-regime model suffices to explain the dynamics (note how regimes B and E are duplicates with high switching frequency between the two).

\subsubsection{Anomaly Detection}
\begin{figure*}
  \centering
  \includegraphics[width=0.95\linewidth]{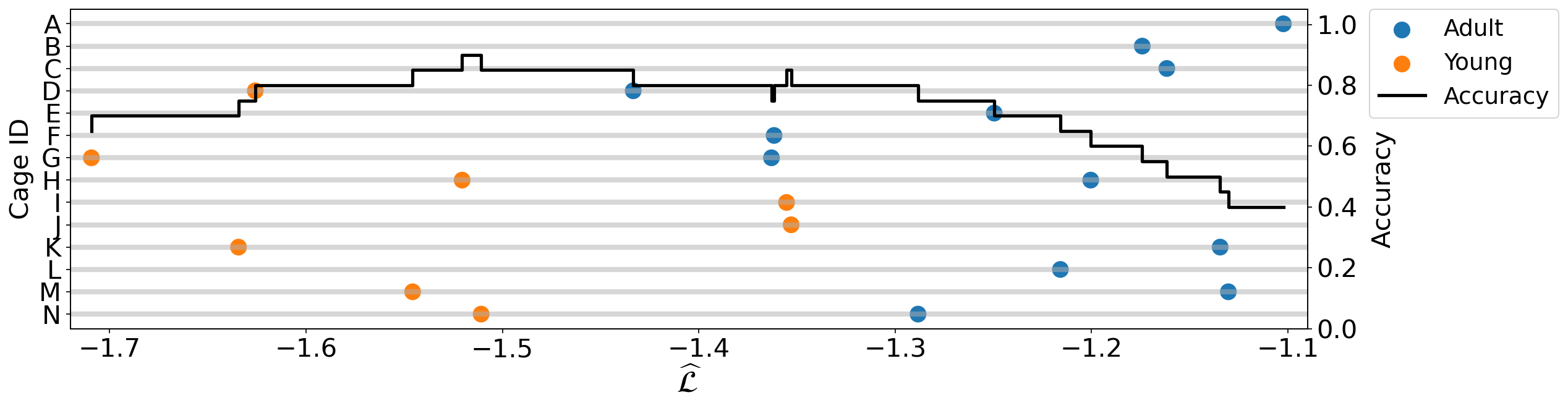}
  \caption{
  	\nll\ scores ($x$-axis) of the \gls{gbm} on each cage ($y$-axis, left) in the adult/young age groups, together with the accuracy of a binary threshold on the \nll\ (scale on the right).
    }
  	\label{FIG_EXP_ABNORMAL}
\end{figure*}

We used the model trained on our `normal' demographic to analyse data from `other' cages: \ie anomaly detection.
This capacity is useful \eg to identify unhealthy mice, strain-related differences, or, as in our proof of concept, evolution of behaviour through age.
In \cref{FIG_EXP_ABNORMAL} we show the trained \gls{gbm} evaluated on data from both the adult (blue) and young (orange) demographics in \gls{imadge}.
Apart from two instances, the \smash{\nll}\ is consistently lower in the younger group compared to the adult demographic: moreover, for all cages where we have data in both age groups, \smash{\nll}\ is always lower for the young mice.
Indeed, a binary threshold achieves 90\% accuracy when optimised and a T-test on the two subsets indicates significant differences ($p$-value $=1.1\times 10^{-4}$).
Given that we used mice from the same strain (indeed even from the same cages), the video recordings are very similar: consequently we expect the \gls{alm} to have similar performance on the younger demographic, suggesting that the differences arise from the behaviour dynamics.

\subsubsection{Analysis of Young mice}
\begin{figure*}
  \centering
  \includegraphics[width=0.95\linewidth]{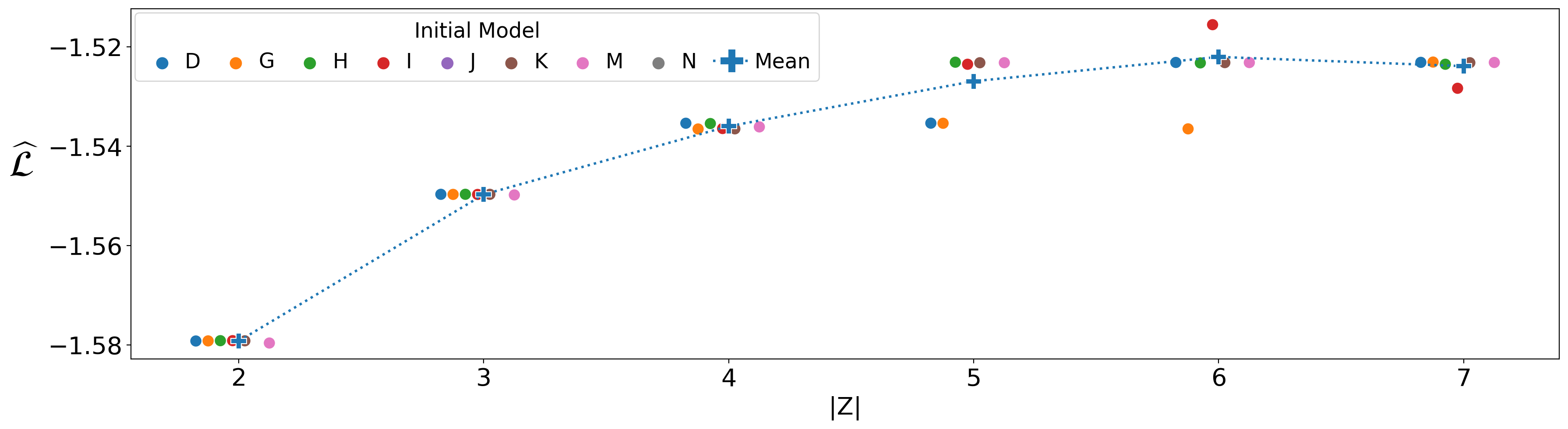}
  \caption{\nll\ as a function of $|Z| \in \{2, 3, ..., 7\}$, with each cage as initialiser. The average (per $|Z|$) is shown as a blue cross.}
  \label{FIG_MDL_YOUNG_NLL_Z}
\end{figure*}

\begin{figure*}
  \centering
  \includegraphics[width=\linewidth]{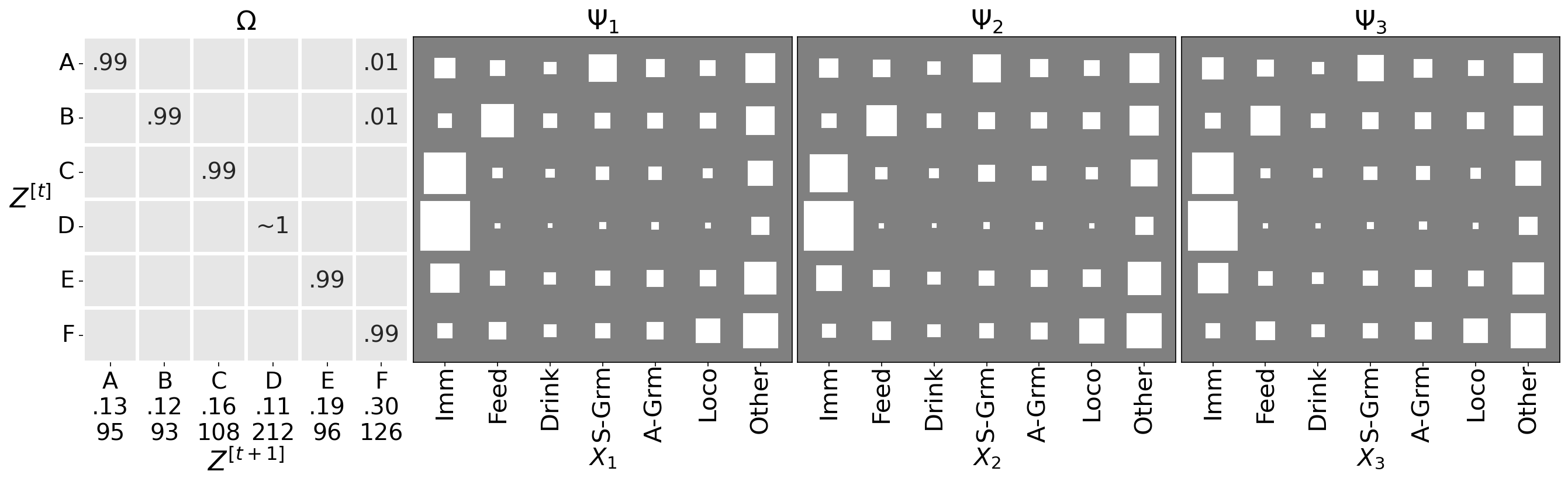}
  \caption{\gls{gbm} parameters on the Young mice data for $|Z|=6$. Arrangement is as in \cref{FIG_MDL_ADULT_PARAMS}.}
  \label{FIG_MDL_YOUNG_PARAMS}
\end{figure*}

Training the model from scratch on the young demographic brings up interesting different patterns.
Firstly, the $|Z|=6$ model emerged as a clear plateau this time, as shown in \cref{FIG_MDL_YOUNG_NLL_Z}.
\Cref{FIG_MDL_YOUNG_PARAMS} shows the parameters for the \gls{gbm} with $|Z|=6$ after optimisation on the young subset.
It is noteworthy that the \IMM state is less pronounced (in
regime D), which is consistent with the younger mice being more
active.
Interestingly, while there is a regime associated with \FEED, it is the same for all mice and also much less pronounced: recall that for the adults, the probability of feeding was 0.7 in each of the \FEED regimes.
This could indicate that the pecking order, at least at the level of feeding, develops with age.

\section{Discussion} \label{S_CONC}
In this paper we have provided a set of tools for biologists to analyse the individual behaviours of group housed mice over extended periods of time.
Our main contribution was the novel \gls{gbm} --- a \gls{hmm} equipped with a permutation matrix for identity matching --- to analyse the joint behaviour dynamics across different cages.
This evidenced interesting dominance relationships, and also flagged significant deviations in an alternative young age group.
In support of the above, we released two datasets, \gls{abode} for training behaviour classifiers and \gls{imadge} for modelling group dynamics (upon which our modelling is based).
\gls{abode} was used to develop and evaluate our proposed \gls{alm} that automatically classifies seven behaviours despite clutter and occlusion.

Since our end-goal was to get a working pipeline to allow us to model the mouse behaviour, the tuning of the \gls{alm} leaves room for further exploration, especially as regards architectures for the \gls{bc}.
In future work we would like to analyse other mouse demographics.
Much of the pipeline should work ``out of the box'', but to handle mice of different colours to those in the current dataset it may be necessary to annotate more data for the \gls{alm} and for the \gls{tim}.

Some contemporary behavioural systems use pose information to determine behaviour.
Note that the \gls{stlt} explicitly uses \gls{bb} poses within the attention architecture of the layout branch: nonetheless, the model was inferior to the \gls{lfb}.
When it comes to limb poses \citet*[sec.\ 5.2]{VL_083} showed that it is very difficult to obtain reliable pose information in our cage setup due to the level of occlusion.
If however, in future work, such pose estimation can be made reliable enough in the cluttered environment of the home-cage, it could aid in improving the classification of some behaviours, such as \SGRM.


\section*{Supplementary Material}
We provide \textbf{Online Resource 1}, a demo video showing the output from the \gls{alm} system on a sample video clip.

\section*{Acknowledgments}
We thank the staff at the \gls{mrc} for providing the raw video and position data, and for their help in interpreting it.
We are also grateful to Andrew Zisserman, for his sage advice on architectures for the behaviour classifier of the \gls{alm}.
Finally, we thank the anonymous reviewers for their comments which have helped to improve the paper.

\section*{Declarations}

\subsection*{Funding}
MPJC was supported by the EPSRC CDT in Data Science (EP/L016427/1).
RSB was supported by funding to MLC from the MRC UK (grant A410).

\subsection*{Ethics Approval}
We emphasize that no new data were collected for this study, in line with the Reduction strategy of the 3Rs~\citep{PHT_022}.
The original observations were carried out at \gls{mrc} in accordance with the Animals (Scientific Procedures) Act 1986, UK, Amendment Regulations 2012 (SI 4 2012/3039).

\subsection*{Availability of data and materials}
The datasets used in this paper were curated by us and have been made publicly available at \urlabode\ and \urlimadge.

\subsection*{Code availability}
Code to produce the above results is available at \CODE.
Note that the correct operation of the STLT and LFB models depends on external repositories outside our responsibility.


\clearpage
\begin{center}
\Huge\textsc{Appendices}
\end{center}
\appendix
\counterwithin{table}{section}
\counterwithin{figure}{section}
\counterwithin{equation}{section}

\section{Derivations}
We defined our \gls{gbm} graphically in \cref{FIG_MDL_GLOBAL_GRAPH} and through \cref{EQ_MDL_GLOBAL_LIKELIHOOD} in the main text.
Herein we derive the update equations for our modified \gls{em} scheme in \cref{ALGO_MDL_GLOBAL_EM}.

\subsection{Notation}
We already defined our key variables in \cref{SS_METHOD_MODELLING} in the main text.
However, in order to facilitate our discussion, we make use of the following additional symbols.
Firstly, let $\mathbf{Q}^{[m]}$ represent the matrix manifestation (outcome in the sample-space) of the random variable ${Q}^{[m]}$.
Secondly, we use $\smash{\mathbb{I}^{[m,n,t]}_k}$ to signify that the observation for mouse $k$ from cage $m$ in sample $t$ of run $n$ is not available: \ie it is missing data.
We assume that this follows a Missing-at-Random mechanism \citep{MISC_011} which allows us to simply ignore such dimensions: \ie $\mathbb{I}$ acts as a multiplier such that it zeros out all entries corresponding to missing observations.

\subsection{Posterior over $Q$}
Due to the deterministic multiplication, selecting a particular $\mathbf{Q}$, and fixing $X$ (because it is observed), completely determines $\smash{\widetilde{X}}$.
Formally:
\begin{equation}
\widetilde{X}^{[m,n,t]} = \left. \mathbf{Q}^{[m]}\right.\Transpose \left(X\mathbb{I}\right)^{[m,n,t]} , \label{EQ_MDL_GLOBAL_XTILDE}
\end{equation}
where we have made use of the fact that for a permutation matrix, the inverse is simply the transpose.
It follows that:
\begin{align}
\Prob{Q^{[m]} = q|X} & \propto \sum_{z',\tilde{x}'}\Prob{q,X,z',\tilde{x}'} \nonumber \\
& \propto \PQ_q \sum_{z'}\Prob{z'}\sum_{\tilde{x}'}\Prob{\tilde{x}'|z'}\Prob{X|\tilde{x}',q} \nonumber \\
& \propto \PQ_q \sum_{z'}\Prob{z'}\Prob{\widetilde{X}_{q}|z'} \nonumber \\
& = \frac{\PQ_q \Prob{\mathbf{Q}\Transpose \left(X\mathbb{I}\right)}}{\sum_{q'}\PQ_q' \Prob{\mathbf{Q'}\Transpose \left(X\mathbb{I}\right)}} , \label{EQ_DERIVE_POST_Q_DEF}
\end{align}
where we made use of the deterministic relationship so that all probabilities over $\tilde{x}$ collapse to 0 if not following the permutation inferred by $q$.
In turn, $\Prob{\mathbf{Q}\Transpose \left(X\mathbb{I}\right)}$ is simply the observed data likelihood of $\smash{\widetilde{X}}$.

\subsection{Complete Likelihood and Auxiliary Function}
Due to \cref{EQ_MDL_GLOBAL_XTILDE}, we can collapse $X$ and $Q$ into $\widetilde{X}$.
Given that we assume the distribution over $Q$ to be sufficiently peaked so that we can pick a single configuration, we can define the complete log-likelihood solely in terms of $Z$ and $\smash{\widetilde{X}}$, much like a \gls{hmm} but with conditionally independent categorical emissions.
Consequently, taking a Bayesian viewpoint and adding priors on each of the parameters, we define the complete data log-likelihood as:
\begin{align}
 \Prob{\SData,\Theta|Q} & = \prod_{m,n}\left( \prod_{z=1}^{|Z|}\PZ_{z}^{Z_{z}^{[m,n,1]}} \prod_{t=2}^{T} \prod_{z',z} \TZ_{z',z}^{Z_{z'}^{[m,n,t-1]}Z_{z}^{[m,n,t]}} \prod_{t,z,k,x}\PX_{k,z,x}^{\widetilde{X}_{k,x}^{[m,n,t]}} \right) \nonumber \\
& \qquad \times \text{Dir}\left(\PZ;\alpha^\PZ\right)\prod_{z=1}^{|Z|}\text{Dir}\left(\TZ_z;\alpha^{\TZ}_z\right) \prod_{k,z}\text{Dir}\left(\PX_{k,z};\alpha^{\PX}_{k,z}\right) , \label{EQ_DERIVE_CHMM_LIKEL}
\end{align}
where 
\begin{equation*}
\text{Dir}\left(\theta;\alpha\right) = \frac{1}{\mathbb{\beta}\left(\alpha\right)}\prod_{i=1}^{|\theta|}\theta_i^{\alpha_i-1}
\end{equation*}
is the usual Dirichlet prior with the multivariate $\beta$ normaliser function for parameter $\theta \in \left\lbrace \PZ, \TZ, \PX \right\rbrace$.
Note that to reduce clutter, we index $\smash{\widetilde{X}}$ using $k$ and $x$ rather than $\tilde{k}/\tilde{x}$.

We seek to maximise the logarithm of the above, but we lack knowledge of the latent regime $Z$.
In its absence, we take the \emph{Expectation} of the log-likelihood with respect to the latest estimate of the parameters ($\hat{\Theta}$) and the observable $\smash{\widetilde{X}}$.
We define this expectation as the \textbf{\textsl{Auxiliary}} function, $\QFunc$ (note the calligraphic form to distinguish from the permutation matrix $Q$):
\begin{align}
\QFunc\left(\Theta, \hat{\Theta}\right) & \equiv \ \mathbb{E}\left\langle \log\left(\Prob{\SData,\Theta|Q}\right)|X, \Theta^{*}\right\rangle \nonumber \\
& = \sum_{m,n} \Biggl( \sum_{z=1}^{|Z|} \mathbb{E}\left\langle Z_{z}^{[m,n,1]}\right\rangle \log\left(\PZ_{z}\right) + \sum_{t=2}^{T} \sum_{z',z} \mathbb{E}\left\langle Z_{z'}^{[m,n,t-1]}Z_{z}^{[m,n,t]} \right\rangle \log\left(\TZ_{z',z}\right) \Biggr) \nonumber \\
& \quad + \sum_{m,n,t,z}\mathbb{E}\left\langle Z^{[m,n,t]}_z \right\rangle \sum_{k,x}\widetilde{X}_{k,x}^{[m,n,t]}\log\left(\PX_{k,z,x}\right) \nonumber \\
& \quad + \; \logprob \label{EQ_DERIVE_HMM_Q}
\end{align}
Note that the number of runs $N$ can vary between cages $m \in M$, and similarly, $T$ is in general different for each run $n$: however, we do not explicitly denote this to reduce clutter.

\subsection{E-Step}
In \cref{EQ_DERIVE_HMM_Q} have two expectations, summarised as:
\begin{align}
\gamma_{z}^{[m,n,t]} \equiv \mathbb{E}\left\langle Z_{z}^{[m,n,t]}\right\rangle = \Prob{Z^{[m,n,t]}=z|\widetilde{X}} \label{EQ_MDL_HMM_GAMMA_COMPUTE}
\end{align}
and
\begin{align}
\eta_{z',z}^{[m,n,t]} \equiv \mathbb{E}\left\langle Z_{z'}^{[m,n,t-1]}Z_{z}^{[m,n,t]} \right\rangle  = \Prob{Z^{[m,n,t-1]} = z', Z^{[m,n,t]} = z | \widetilde{X}} \label{EQ_MDL_HMM_ETA_COMPUTE} .
\end{align}
The challenge in computing these is that it involves summing out all the other $z* \notin \left\lbrace z, z'\right\rbrace$.
This can be done efficiently using the recursive updates of the Baum-Welch algorithm \citep{ND_089}, which is standard for \glspl{hmm}.

\subsubsection{Recursive Updates}
We first split the dependence around the point of interest $t$.
To reduce clutter, we represent indexing over $m/n$ by `$.$' on the right hand side of equations and summarise the emission probabilities as:
\begin{equation*}
P_{\widetilde{X}}^{[m,n]}\left(t,z\right) \equiv \left(\prod_{k=1}^K\prod_{x=1}^{|\widetilde{X}|}\PX_{k,z,x}^{\widetilde{X}_{k,x}^{[.,t]}}\right)
\end{equation*}
Starting with $\gamma$:
\begin{align}
\gamma_z^{[m,n,t]} & = \frac{\Prob{\widetilde{X}|Z^{[.,t]}_{z}}\Prob{Z^{[.,t]}_{z}}}{\Prob{\widetilde{X}}} \nonumber \\
& = \frac{\Prob{\widetilde{X}^{[.,1:t]}|Z^{[.,t]}_{z}} \Prob{\widetilde{X}^{[.,t+1:T]}|Z^{[.,t]}_{z}} \Prob{Z^{[.,t]}_{z}}}{\Prob{\widetilde{X}}} \nonumber \\
& = \frac{\Prob{\widetilde{X}^{[.,1:t]},Z^{[.,t]}_{z}} \Prob{\widetilde{X}^{[.,t+1:T]}|Z^{[.,t]}_{z}} }{\Prob{\widetilde{X}}} . \label{EQ_DERIVE_HMM_GAMMA_EXPAND}
\end{align}
Similarly, for $\eta$:
\begin{align}
\eta_{z',z}^{[m,n,t]} & = \frac{\Prob{\widetilde{X} | Z^{[.,t-1]}_{z'}, Z^{[.,t]}_{z}} \Prob{Z^{[.,t-1]}_{z'}, Z^{[.,t]}_{z}}}{\Prob{\widetilde{X}}} \nonumber \\
&  = \frac{\Biggl( \Prob{\widetilde{X}^{[.,1:t-1]} | Z^{[.,t-1]}_{z'}} P_{\widetilde{X}}^{[.]}\left(t,z\right) \Prob{\widetilde{X}^{[.,t+1:T]} | Z^{[.,t]}_{z}}\Prob{Z^{[.,t-1]}_{z'}} \TZ_{z',z} \Biggr)}{\Prob{\widetilde{X}}} \nonumber \\
& = \frac{\Biggl( \Prob{\widetilde{X}^{[.,1:t-1]}, Z^{[.,t-1]}_{z'}} P_{\widetilde{X}}^{[.]}\left(t,z\right) \Prob{\widetilde{X}^{[.,t+1:T]} | Z^{[.,t]}_{z}} \TZ_{z',z} \Biggr)}{\Prob{\widetilde{X}}} \label{EQ_DERIVE_HMM_ETA_EXPAND}
\end{align}
We see that now we have two `messages' that crucially can be defined recursively.
Let the `forward' pass\footnote{In some texts these are usually referred to as $\alpha$ and $\beta$ but we use $F/B$ to avoid confusion with the parameters of the Dirichlet priors.} be denoted by $F$ as:
\begin{align}
F^{[m,n,t]}_{z} & = \Prob{\widetilde{X}^{[.,1:t]}, Z^{[.,t]}_{z}} \nonumber \\
&  = \sum_{z'=1}^{|Z|} \Biggl( \Prob{\widetilde{X}^{[.,1:t-1]}, Z^{[.,t-1]}_{z'}} \Prob{Z^{[.,t]}_{z} | Z^{[.,t-1]}_{z'}} \Prob{\widetilde{X}^{[.,t]} | Z^{[.,t]}_{z}} \Biggr) \nonumber \\
& = P_{\widetilde{X}}^{[.]}\left(t,z\right) \sum_{z'=1}^{|Z|} F^{[.,t-1]}_{z'} \TZ_{z',z} \nonumber .
\end{align}
For the special case of $t=1$, we have:
\begin{equation*}
F^{[m,n,1]}_{z} = \PZ_{z} P_{\widetilde{X}}^{[.]}\left(1,z\right) .
\end{equation*}
Similarly, we denote the `backward' recursion by $B$:
\begin{align}
B^{[m,n,t]}_{z} & = \Prob{\widetilde{X}^{[.,t+1:T]}|Z^{[.,t]}_{z}} \nonumber \\
	& = \sum_{z'=1}^{|Z|}\Biggl(\Prob{Z^{[.,t+1]}_{z'}|Z^{[.,t]}_{z}} \Prob{\widetilde{X}^{[.,t+1]}|Z^{[.,t+1]}_{z'}} \Prob{\widetilde{X}^{[.,t+2:T]}|Z^{[.,t+1]}_{z'}} \Biggr) \nonumber \\
	& = \sum_{z'=1}^{|Z|} \TZ_{z,z'} P_{\widetilde{X}}^{[.]}\left(t+1,z\right) B^{[.,t+1]}_{z'} \nonumber .
\end{align}
Again, we have to consider the special case for $t=T$:
\begin{equation*}
B^{[m,n,T]}_{z} = 1 .
\end{equation*}

\subsubsection*{Scaling Factors}
To avoid numerical underflow, we work with normalised distributions.
Specifically, we define:
\begin{equation*}
\hat{F}^{[m,n,t]}_z = \Prob{Z^{[.,t]}_z|\widetilde{X}^{[.,1:t]}} = \frac{F^{[.,t]}_z}{\Prob{\widetilde{X}^{[.,1:t]}}} .
\end{equation*}
We relate these factors together through:
\begin{equation*}
S^{[m,n,t]} = \Prob{\widetilde{X}^{[.,t]}|\widetilde{X}^{[.,1:t-1]}} ,
\end{equation*}
and hence, from the product rule, we also have:
\begin{equation*}
\Prob{\widetilde{X}^{[m,n,1:t]}} = \prod_{\tau=1}^{t}S^{[.,\tau]} .
\end{equation*}
Consequently, we can redefine:
\begin{align*}
\hat{F}^{[m,n,t]}_z & = \frac{F^{[.,t]}_z}{\prod_{\tau=1}^{t}S^{[.,\tau]}} \\
\intertext{and}
\hat{B}^{[m,n,t]}_z & = \frac{B^{[.,t]}_z}{\prod_{\tau=t+1}^{T}S^{[.,\tau]}}
\end{align*}
We denote for simplicity
\begin{equation*}
C^{[m,n,t]} = \left(S^{[.,t]}\right)^{-1}
\end{equation*}
as the normaliser for the probability.
This allows us to redefine the recursive updates for the responsibilities as follows:
\begin{align}
\gamma_{z}^{[m,n,t]} & = \hat{F}^{[.,t]}_{z}\hat{B}^{[.,t]}_{z} , \label{EQ_DERIVE_HMM_GAMMA_RECURSE} \\
\intertext{and}
\eta_{z',z}^{[m,n,t]} & = C^{[.,t]}\hat{F}^{[.,t-1]}_{z'} \hat{B}_{z}^{[.,t]} \TZ_{z',z} P_{\widetilde{X}}^{[.]}\left(t,z\right) \label{EQ_DERIVE_HMM_ETA_RECURSE},
\end{align}
where:
\begin{align*}
\hat{F}^{[m,n,t]}_{z} & = C^{[.,t]}\ddot{F}^{[.,t]}_{z} \\
\intertext{and:}
\ddot{F}^{[m.n,t]}_{z} & = P_{\widetilde{X}}^{[.]}\left(t,z\right) \sum_{z'=1}^{|Z|} \hat{F}^{[.,t-1]}_{z'} \TZ_{z',z} \\
\intertext{when $t > 1$, and:}
\ddot{F}^{[m.n,t]}_{z} & = P_{\widetilde{X}}^{[.]}\left(1,z\right) \PZ_{z} \\
\intertext{if $t = 1$. Reformulating:}
C^{[m,n,t]} & = \left(\sum_{z'=1}^{|Z|} \ddot{F}^{[.,t]}_{z'}\right)^{-1} \\
\intertext{with:}
\hat{B}^{[m,n,t]}_{z} & = C^{[.,t+1]}\sum_{z'} \TZ_{z,z'} P_{\widetilde{X}}^{[.]}\left(t+1,z\right) \hat{B}^{[.,t+1]}_{z'}
\intertext{for the general case ($t<T$) and:}
\hat{B}^{[m,n,t]}_{z} & = 1
\end{align*}
when $t = T$.
Through the normalisers $C$, we also compute the observed data log-likelihood:
\begin{equation}
\log\left(\Prob{\widetilde{X};\Theta}\right) = - \sum_{m=1}^M\sum_{n=1}^N\sum_{t=1}^{T} \log\left[C^{[.,t]}\right] . \label{EQ_DERIVE_HMM_OBS_LL}
\end{equation}

\subsection{M-Step}
We re-arrange the $\QFunc$-function to expand and split all terms according to the parameter involved (to reduce clutter we collapse the sum over $M/N$ and ignore constant terms):
\begin{align*}
\QFunc\left(\Theta,\hat{\Theta}\right) & = \sum_{m,n}\sum_{z=1}^{|Z|} \gamma_{z}^{[.,1]} \log\left(\PZ_{z}\right) + \sum_{z=1}^{|Z|} \left(\alpha^{\PZ}_{z} - 1\right) \log\left(\PZ_{z}\right) + \sum_{m,n}\sum_{t=2}^{T} \sum_{z',z} \eta_{z',z}^{[.,t]}\log\left(\TZ_{z',z}\right) \nonumber \\
& \quad + \sum_{z',z} \left(\alpha^{\TZ}_{z',z} - 1\right) \log\left(\TZ_{z',z}\right) + \sum_{m,n}\sum_{t=1}^{T}\sum_{z=1}^{|Z|}\gamma^{[.,t]}_{z}\sum_{k=1}^K \sum_{x=1}^{|\widetilde{X}|}{\widetilde{X}^{[.,t]}_{k,x}}\log\left(\PX_{k,z,x}\right) \nonumber \\
& \quad + \sum_{z=1}^{|Z|}\sum_{k=1}^K\sum_{x=1}^{|\widetilde{X}|}\left(\alpha^{\PX}_{k,z,x}-1\right)\log\left(\PX_{k,z,x}  \right) + Const
\end{align*}

\subsubsection{Maximising for \PZ}
Since we have a constraint (\PZ\ must be a valid probability that sums to 1) we maximise the constrained Lagrangian:
\begin{equation*}
\Lambda = \QFunc + \lambda\left(\sum_{z'=1}^{|Z|}\PZ_{z'} - 1\right) .
\end{equation*}
We maximise this by taking the derivative with respect to $\PZ_z$ and setting it to 0 (note that we can zero-out all terms involving $z' \neq z$ which are constant with respect to $\PZ_z$):
\begin{align}
\frac{\partial\Lambda}{\partial\PZ_z} & = \frac{1}{\PZ_z}\left(\sum_{m=1}^{M}\sum_{n=1}^N \gamma^{[m,n,1]}_{z} + \alpha^\PZ_z-1 \right) + \lambda = 0 \\
\lambda\PZ_z & = - \sum_{m=1}^M\sum_{n=1}^N \gamma^{[m,n,1]}_{z} - \alpha^\PZ_z + 1 \label{EQ_DERIVE_MOC_M_PZ_S1}
\end{align}
Summing the above over $z$:
\begin{equation}
\lambda = - \sum_{m=1}^M\sum_{n=1}^N\sum_{z'=1}^{|Z|} \gamma^{[m,n,1]}_{z'} - \sum_{z'=1}^{|Z|}\alpha^\PZ_{z'} + |Z| = - \sum_{m=1}^{M}N^{m} - \sum_{z'=1}^{|Z|}\alpha^\PZ_{z'} + |Z| \label{EQ_DERIVE_MOC_M_PZ_S2}
\end{equation}
In the above we have made use of the fact that both $\PZ_z$ and $\gamma_z^{[n]}$ sum to 1 over $z'$.
Substituting Eq.\ \eqref{EQ_DERIVE_MOC_M_PZ_S2} for $\lambda$ in Eq.\ \eqref{EQ_DERIVE_MOC_M_PZ_S1} we get the maximum-a-posteriori estimate for $\hat{\PZ}_z$:
\begin{equation}
\hat\PZ_z = \frac{\sum_{m=1}^M\sum_{n=1}^N\gamma^{[m,n,1]}_{z} + \alpha^\PZ_{z} - 1} {\sum_{m=1}^{M}N^{m} + \sum_{z'=1}^{|Z|}\alpha^\PZ_{z'} - |Z|} . \label{EQ_MOC_M_PI_MLE}
\end{equation}

\subsubsection{Maximising for \PX}
We follow a similar constrained optimisation procedure for \PX, with the Lagrangian:
\begin{equation}
\Lambda = \QFunc + \sum_{k', z'}\lambda_{k',z'}\left(\sum_{x'}\PX_{k',z',x'} - 1\right) \label{EQ_DERIVE_MOC_M_PX_S0}
\end{equation}
Taking the derivative of Eq.\ \eqref{EQ_DERIVE_MOC_M_PX_S0} with respect to $\PX_{k,z,x}$ and setting it to 0 (ignoring constant terms):
\begin{align}
\frac{\partial \Lambda}{\partial\PX_{k,z,x}}  = \frac{1}{\PX_{k,z,x}} \left(\sum_{m,n,t}\gamma^{[m,n,t]}_{z}\widetilde{X}^{[m,n,t]}_{k,x} + \alpha^\PX_{k,z,x} - 1 \right) + \lambda_{k,z}\nonumber
\end{align}
Setting this to 0:
\begin{align}
& \lambda_{k, z}\PX_{k,z,x} = - \sum_{m,n,t}\gamma^{[m,n,t]}_{z}\widetilde{X}^{[m,n,t]}_{k,x} - \alpha^\PX_{k,z,x} + 1 \label{EQ_DERIVE_MOC_M_PX_S1} \\
\intertext{Again, summing this over $x'$ yields:}
& \lambda_{k,z} = - \sum_{m,n,t}\gamma^{[m,n,t]}_{z}\sum_{x'}\widetilde{X}^{[m,n,t]}_{k,x'} - \sum_{x'}\alpha^\PX_{k,z,x'} + |\widetilde{X}|  \label{EQ_DERIVE_MOC_M_PX_S2}
\end{align}
Substituting Eq.\ \eqref{EQ_DERIVE_MOC_M_PX_S2} back into Eq.\ \eqref{EQ_DERIVE_MOC_M_PX_S1} gives us:
\begin{align}
\hat{\PX}_{k,z,x} = \frac{\sum_{m,n,t}\gamma^{[m,n,t]}_{z}\widetilde{X}^{[m,n,t]}_{k,x} + \alpha^\PX_{k,z,x} - 1}{\sum_{m,n,t}\gamma^{[m,n,t]}_{z}\sum_{x'}\widetilde{X}^{[m,n,t]}_{k,x'} + \sum_{x'}\alpha^\PX_{k,z,x'} - |\widetilde{X}|} . \label{EQ_MOC_M_PSI_MLE}
\end{align}

\subsubsection{Maximising for \TZ}
As always, this is a constrained optimisation by virtue of the need for valid probabilities.
We start from the Lagrangian:
\begin{align}
& \Lambda = \QFunc + \sum_{z^\dagger=1}^{|Z|}\lambda_{z^\dagger}\left( \sum_{z^*=1}^{|Z|}\TZ_{z^\dagger,z^*} - 1 \right) \nonumber \\
& \frac{\partial \Lambda}{\partial\TZ_{z',z}} = \frac{1}{\TZ_{z',z}}\left( \sum_{m,n} \sum_{t=2}^{T} \eta_{z',z}^{[., t]}  + \left(\alpha^\TZ_{z',z}-1\right) \right) + \lambda_{z'} \nonumber \\
& \lambda_{z'}\TZ_{z',z} = - \left( \sum_{m,n} \sum_{t=2}^{T} \eta_{z',z}^{[., t]}  + \left(\alpha^\TZ_{z',z}-1\right) \right) \nonumber \\
& \lambda_{z'} = - \left( \sum_{m,n} \sum_{t=2}^{T} \sum_{z^*=1}^{|Z|}\eta_{z',z^*}^{[., t]}  + \sum_{z^*=1}^{|Z|}\alpha^\TZ_{z',z^*} - |Z| \right) \nonumber
\end{align}
which after incorporating into the previous equation gives the maximum-a-posteriori update:
\begin{equation}
\hat{\TZ}_{z',z} = \frac{\sum_{m,n} \sum_{t=2}^{T} \eta_{z',z}^{[., t]}  + \left(\alpha^\TZ_{z',z}-1\right)}{\sum_{m,n} \sum_{t=2}^{T} \sum_{z^*}\eta_{z',z^*}^{[., t]}  + \sum_{z^*}\alpha^\TZ_{z',z^*} - |Z|} \label{EQ_MDL_HMM_TZ_UPDATE}
\end{equation}


\section{Labelling Behaviour in \gls{abode}}\label{APP_ABODE}
A significant effort in the curation of the \gls{abode} data was the annotation of the behaviours.
This involved development of a well-defined annotation schema, and a rigorous annotation process, building upon the expertise of the animal care technicians at \gls{mrc} and our own experience in annotation processes.

\subsection{Annotation Schema}\label{SS_APP_ANNOTATION}
Labels are specified per-mouse per-\gls{btp}, focusing on both the \emph{behaviour} and also an indication of whether it is \emph{observable}.

\subsubsection{Behaviour}
The schema admits nine behaviours and three other labels, as shown in \cref{TAB_DATA_SCHEMA_BEHAVIOUR}.
In particular, labels \HID, \NOID, \TENT and \OTR ensure that the annotator can specify a label in every instance, and clarify the source of any ambiguity.
In this way, the labels are mutually exclusive and exhaustive.
This supports our desire to ensure that each \gls{btp} is given exactly one behaviour label and eliminates ambiguity about the intention of the annotator.

\begin{table*}
	\centering
	\setlength\tabcolsep{1.2mm}
	\caption{
	Definition of Behaviour Labels: these are listed in order of precedence (most important ones first).
	The short-hand label (used in figures/tables) is in square brackets.
	}
	\label{TAB_DATA_SCHEMA_BEHAVIOUR}
	\begin{tabular}{p{2.5cm}p{13cm}}
\toprule[1.5pt]
\textbf{Behaviour} & \textbf{Description}\\
\midrule
\HID \newline [Hid] & Mouse is fully (or almost fully) occluded and barely visible. Note that while the annotator may have their own intuition of what the mouse is doing (because they saw it before) they should still annotate as Hidden. \\[0.5em]
\NOID [N/ID] & Annotator cannot identify the mouse with certainty. Typically, there is at least another mouse which has an Unidentifiable flag. \\[0.5em]
\IMM \newline [Imm] & Mouse is not moving and static (apart from breathing), which may or may not be sleeping. \\[0.5em]
\FEED \newline [Feed] & The mouse is eating, typically with its mouth/head in the hopper: it may also be eating from the ground. \\[0.5em]
\DRINK \newline [Drink] & Drinking from the water spout. \\[1.5em]
 \SGRM \newline [S-Grm] & Mouse is grooming itself. \\[1.5em]
 \AGRM [A-Grm] & Mouse is grooming another cage member. In this case, the annotator \textbf{must} indicate the recipient of the grooming through the Modifier field. \\[0.5em]
\CLIMB [Climb] & All feet off the floor and also NOT on the tunnel, with the nose outside the food hopper if it is using it for support (\ie it should not be eating as well). \\[0.5em]
\UMVE [uMove] & A general motion activity while staying in the same place. This could include for example sniffing/looking around/rearing. \\[0.5em]
\LOCO \newline [Loco] & Moving/Running around. \\[1.5em]
\TENT [Tent] & The mouse is exhibiting one of the behaviours in the schema, but the annotator is uncertain of which. If possible, the subset of behaviours that are tentative should be specified as a Modifier. In general, this is an indication that the behaviour needs to be evaluated by another annotator. \\[0.5em]
\OTR \newline [Other] & Mouse is doing something which is not accounted for in this schema. Certainly, aggressive behaviour will fall here, but there may be other behaviours we have not considered. \\
\bottomrule[1.5pt]
\end{tabular}
\end{table*}

\subsubsection{Observability}
The \HID label, while treated as mutually exclusive with respect to the other behaviours for the purpose of annotation, actually represents a hierarchical label space.
Technically, \HID mice are doing any of the other behaviours, but we cannot tell which --- any subsequent modelling might benefit from treating these differently.
We thus sought to further specify the \textsl{observability} of the mice as a label-space in its own right as shown in \cref{TAB_DATA_SCHEMA_OBSERVABILITY}.

\begin{table*}
	\centering
	\caption{Definitions of Observability labels (shorthand label in square brackets)}
	\label{TAB_DATA_SCHEMA_OBSERVABILITY}
	\setlength\tabcolsep{1.2mm}
	\begin{tabular}{p{2.8cm}p{12.5cm}}
\toprule[1.5pt]
\textbf{Observability} & \textbf{Description}\\
\midrule
\NOOB \newline [N/Obs] & Mouse is fully (or almost) occluded and not enough information to give any behaviour. When mice are huddling (and clearly immobile), a mouse is still considered Not Observable if none of it is visible. \\[0.5em]
\OBS \newline [Obs] & Mouse is visible (or enough to distinguish between some behaviours). Note that it may still be difficult to identify with certainty what the mouse is doing but at a minimum can differentiate between `Immobile' and other behaviours. \\[0.5em]
\AMB \newline [Amb] & All other cases, especially when it is borderline or when mouse cannot be identified with certainty.\\
\bottomrule[1.5pt]
\end{tabular}
\end{table*}

\subsection{Annotation Process}
The annotations were carried out using the BORIS software \citep{MISC_049}: this was chosen for its versatility, familiarity to the authors and open-source implementation.

\subsubsection{Recruitment}
Given the resource constraints in the project, we were only able to recruit a single expert animal care technician (henceforth the \emph{\PHENO}) to do our annotations.
This limited the scale of our dataset but on the other hand simplified the data curation process and ensured consistency throughout the dataset.
In particular, it should be noted that given the difficulty of the task (which requires very specific expertise), the annotation cannot be crowdsourced.
\citet[Sec.\ 3.1.4]{MISC_060} documents attempts to use behaviour annotations obtained through an online crowdsourcing platform: initial in-depth analysis of the data indicated that the labellings were not of sufficiently high quality as to be reliable for training our models.
To mitigate the potential shortcomings of a single annotator, we:
(a) carried out a short training phase for the \PHENO with a set of clips that were simultaneously annotated by the \PHENO and ourselves,
(b) designed some automated sanity checks to be run on annotations (see below), \textsl{and},
(c) re-annotated the observability labels ourselves.

\subsubsection{Method}
The \PHENO was provided with the 200 two-minute clips, grouped in batches of 20 clips each (10 batches in total, 400 minutes of annotations).
The clips were randomised and stratified such that in each batch there are 10 clips from the training split, 4 clips from the validation split and 6 from the testing split.
This procedure was followed to reduce the effect of any shift in the annotation quality (as the \PHENO saw more of the data) on the datasplits.

For each clip, the \PHENO had access to the CLAHE-processed video and the \gls{rfid}-based position information.
We made the conscious decision to not use the \gls{bb} localisation from the \gls{tim} since this allows us to (a) decouple the behaviour annotations from the performance of upstream components, and (b) provides a more realistic estimate of end-to-end performance.

Although behaviour is defined per \gls{btp}, annotating in this manner is not efficient for humans: instead, the annotator was tasked with specifying intervals of the specific behaviour, defined by the start and end-point respectively.
This is also the motivation behind limiting clips to two-minute snippets: these are long enough that they encompass several behaviours but are more manageable than the 30-minute segments, and also provide more variability (as it allows us to sample from more cages).

\subsubsection{Quality Control}
To train the \PHENO, we provided a batch of four (manually chosen) snippets, which were also annotated by ourselves --- this enabled the \PHENO to be inducted into using BORIS and in navigating the annotation schema, providing feedback as required.
Following the annotation of each production batch, we also ran the labellings through a set of automated checks which guarded against some common errors.
These were reported back to the \PHENO, although they had very limited time to act on the feedback which impacted on the resulting data quality.

The main data quality issue related to the misinterpretation of \HID by the \PHENO, leading to over-use of the \HID label.
To rectify this, we undertook to re-visit the samples labelled as \HID and clarify the observability as per the schema in \cref{TAB_DATA_SCHEMA_OBSERVABILITY}.
Samples which the \PHENO had labelled as anything other than \HID (except for \NOID samples which were ignored as ambiguous) were retained as \OBS --- we have no reason to believe that the \PHENO reported a behaviour when it should have been \HID.
The only exception was when there was a clear misidentification of the mice, which was rectified (we had access to the entire segment which provided longer-term identity cues for ambiguous conditions).
Note that our annotation relates to the observability (or otherwise): however, when converting a previously \HID sample to \OBS, we provided a ``\emph{best-guess}'' annotation of behaviour.
These best-guess annotations are clearly marked, allowing us to defer to the superior expertise of the \PHENO in differentiating between actual behaviours where this is critical (\eg for training models).

\subsubsection{Statistics}
We end this section by reporting dataset statistics.
\gls{abode} consists of 200 snippets, each two-minutes in length.
These were split (across recording boundaries) into a Training, Validation and Test set.
\Cref{TAB_ABODE_STATS} shows the number of admissible mouse-\glspl{btp} that are annotated observable/hidden and the distribution of behaviours for observable samples.

\begin{table}[!h]
\centering
\setlength\tabcolsep{1.5mm}
\caption{Statistics for \gls{abode}, showing the number of samples (\glspl{btp}-mouse pairs) for each behaviour and data partition.}
\label{TAB_ABODE_STATS}
\begin{tabular}{llrrr}
\toprule[1.5pt]
          &  &   Train &  Validate &  Test \\
\midrule
Visibility & Observable &  35297 &      9374 &  19075 \\
          & Hidden &   2650 &       750 &   1506 \\
Behaviour & Imm &  19363 &      5455 &  10462 \\
          & Feed &   3298 &       750 &   2314 \\
          & Drink &    272 &        72 &    161 \\
          & S-Grm &   2670 &       800 &   1512 \\
          & A-Grm &   1278 &       339 &    550 \\
          & Loco &    959 &       177 &    375 \\
          & Other &   7457 &      1781 &   3701 \\
\bottomrule[1.5pt]
\end{tabular}
\end{table}

\clearpage
\bibliographystyle{IEEETransN}

\end{document}